\documentclass[letterpaper, 10 pt, conference]{ieeeconf}

\usepackage{mathtools}
\usepackage{amsmath,mleftright}
\usepackage{xparse}
\usepackage{comment}
\usepackage{amssymb}

\usepackage{amsthm}
\usepackage{graphicx}
\usepackage{amsmath}
\usepackage{amssymb}
\usepackage{subfig}
\let\labelindent\relax
\usepackage{enumitem}
\PassOptionsToPackage{noadjust}{cite} 

\usepackage{cite}
\usepackage{tabularx}
\usepackage[table]{xcolor}
\usepackage[colorlinks=true,urlcolor=blue]{hyperref}  
\usepackage[noadjust]{cite}
\usepackage{subcaption}
\usepackage{subfig}

\newtheorem{lemma}{Lemma}%

\newtheorem{proposition}{Proposition}

\newtheorem{definition}{Definition}%

\hypersetup{
    colorlinks=true,
    linkcolor=darkgray,
    filecolor=magenta,      
    urlcolor=cyan,
    pdftitle={Overleaf Example},
    pdfpagemode=FullScreen,
     citecolor    = darkgray
    }

 	\definecolor{newgreen}{RGB}{0,127,0}
  \definecolor{newblue}{RGB}{0,0,127}

\IEEEoverridecommandlockouts

\newcommand{\Ee}{\mathbb{E}_{\mathcal{C}}^{r}}

\usepackage{titlesec}

\usepackage{xcolor} 

\begin{document}

\title{Constructive Infinitely Differentiable Guidance Vector Fields for Robotic Control}

\author{
    Vinicius Mariano Gonçalves, Peilin Wu,
    Adriano Rezende, and Weijia Yao%
    \thanks{Vinicius Mariano Gonçalves is with Federal University of Minas Gerais (UFMG), Brazil. Adriano Rezende is independent. }%
    \thanks{Peilin Wu and Weijia Yao are with the School of
Robotics, Hunan University, Changsha, China.}%
}

\maketitle

\begin{abstract}
Guidance vector fields (GVFs) are a popular framework for tasks such as robotic path-following and multi-robot motion coordination, but existing approaches may be non-differentiable or have large derivatives, limiting their use in controllers that require derivatives of the guidance signal, which is common when the GVF is used as an input in a cascaded controller. This paper proposes an infinitely differentiable constructive GVF. The resulting vector field depends on a smoothing parameter and is a generalization of \cite{AdrianoIEEE}, recovering the original GVF in the limit as this parameter approaches zero. It also admits a simple and efficient implementation that preserves smoothness under discretization and can reduce computation time by using substantially fewer curve samples while maintaining small path-following errors. We establish the main theoretical properties and demonstrate its practical advantages through numerical and real-world experiments.
\end{abstract}

\section{Introduction}

Guidance vector fields (GVFs) have been widely used for robot control, particularly for path following, owing to their conceptual simplicity and strong theoretical foundations \cite{goncalves2010,AdrianoIEEE,yao2021singularity,yao2022topological}. More recent developments include nonsingular constructions, multi-robot coordination, inverse-kinematics formulations, data-based representations, extensions to Lie groups, and safety-aware formulations \cite{chen2025nonsingular,zhou2025inverse,chen2025discrete,bartelt2026constructive,nunes2026safe,yao2022guiding}. Given a static target curve in the configuration space, a GVF usually assigns to each configuration in its domain a desired velocity vector. Following this velocity field causes the system to approach and traverse the target curve. Among these approaches, \cite{AdrianoIEEE} proposed a practical constructive GVF for generic curves with formal convergence guarantees. Its construction is based on the Euclidean distance and the closest-point map between the current configuration and the target curve.

In many robotic systems, however, the velocity generated by the GVF is not the physical control input. Instead, this desired velocity must be realized by a lower-level controller acting through accelerations, forces, or torques. A common solution is to employ a cascade architecture, for example through a backstepping design \cite{nunes2023}, in which the velocity generated by the GVF is supplied as a reference to a dynamics-level control layer. In a quadrotor, for instance, the propeller thrusts and the resulting torques can be controlled so that the vehicle tracks the desired linear velocity prescribed by the GVF.

Implementing such a cascade architecture often requires time derivatives of the GVF output \cite{nunes2023}. In some designs, derivatives of order higher than one are required; for example, differential-flatness-based quadrotor controllers may use the acceleration, jerk, and snap induced by the desired velocity field \cite{zhou2014vector}. These derivatives appear in feedforward acceleration, force, or torque terms that are important both in formal stability analyses and in improving practical tracking performance \cite{zhou2014vector,nunes2023,gao2025distributed}.

\begin{figure}[t]
    \centering
    \includegraphics[width=1\linewidth]{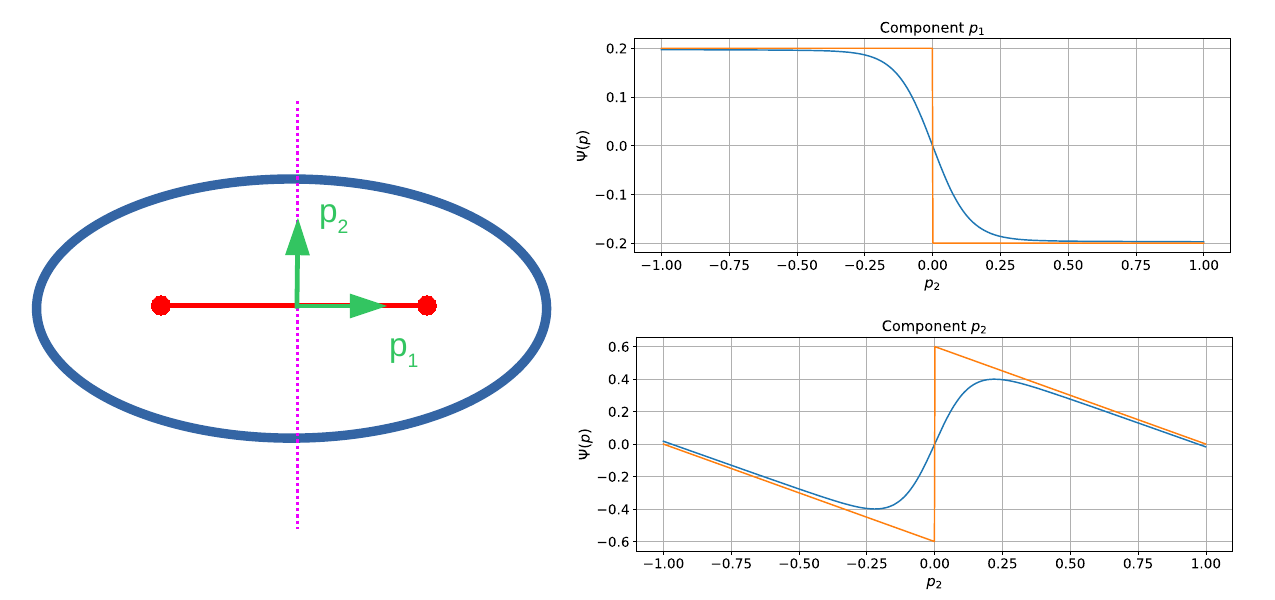}
    \caption{Illustration of the discontinuity problem in \cite{AdrianoIEEE}. The target curve is the ellipse, and the red line segment represents its medial axis, where the distance function is nondifferentiable. The plots on the right show the two components of the GVF in \cite{AdrianoIEEE} (orange) and of the proposed GVF (blue), evaluated along the magenta line. Note the discontinuity of the orange vector field at $p_2=0$, which lies on the medial axis.}
    \label{fig:vector_field}
\end{figure}

The required derivatives, however, may fail to exist. The construction in \cite{AdrianoIEEE}, for example, relies on the Euclidean distance between the configuration and the curve and its derivative, which is related to the closest point between the current point and the curve. Although the Euclidean distance is continuous everywhere, it is generally non-differentiable at points having multiple closest points on the curve, a set called the \emph{medial axis}. At the center of a circle, for instance, every point on the circle is a closest point. At such configurations, the GVF is discontinuous. See Figure \ref{fig:vector_field} for an illustration of this phenomenon.

These singular configurations  form a set of measure zero and may not be reachable by trajectories initialized outside it. Nevertheless, they can still create practical difficulties. Near the medial axis, the vector field can change very rapidly. Consequently, a small perturbation in the robot configuration may produce a large change in the desired velocity and, therefore, in the feedforward control input generated from its derivatives.

A second source of non-smoothness, perhaps even more important in practice, arises from the numerical implementation of the closest-point computation. Except for very simple curves, the Euclidean distance and closest point cannot be obtained analytically. One possible implementation is to discretize the curve into a finite collection of samples and select the sample closest to the current configuration. This procedure introduces a hard nearest-neighbor decision. As the configuration crosses a boundary between the Voronoi regions of two samples, the selected sample may change abruptly, producing a non-smooth approximation of the GVF even in regions where the original GVF was smooth. Numerical differentiation of this approximate field can then generate very large values. Unlike the intrinsic medial axis singularity, this phenomenon may occur even arbitrarily close to the target curve, which is particularly concerning because a path-following robot is expected to operate close to the curve during most of its motion.

In this paper, we propose a  construction which is a generalization of \cite{AdrianoIEEE} when we consider static curves. The resulting GVF is parameterized by a positive scalar $r$, which controls the trade-off between tracking precisely the original target curve and  having smoother control inputs. For every $r>0$, the proposed vector field is infinitely differentiable at every point of the configuration space. This property is preserved when the defining integrals are replaced by a fixed numerical quadrature over sampled points of the curve, which is straightforward to implement. As $r\rightarrow +0$, the proposed vector field converges to the original vector field of \cite{AdrianoIEEE} at configurations for which the closest-point evaluation operation is well defined. 

The construction is based on what we call the \emph{$r$-Differentiable Evaluation Operator} ($r$-DEO). The operator replaces evaluation of a function at a closest point on the curve with a normalized Gaussian-weighted average over the entire curve. It also admits a probabilistic interpretation. The normalized exponential weights used by the DEO are closely related to those arising from \emph{log-sum-exp} and \emph{soft-min} approximations of hard maximum and minimum operators \cite{blanchard2021logsumexp}. Log-sum-exp regularization has been used in several control applications to replace non-smooth objects with differentiable approximations. In model predictive control, smooth maximum approximations have been employed to obtain differentiable soft penalties \cite{RICHARDS201551}. Smooth approximations of geometric distances have also been used together with control barrier functions to produce smoother distance-dependent safety constraints and control signals \cite{Goncalves2024SafeSmooth}. Similarly, soft-minimum and soft-maximum operators have been used to regularize minimum-based barrier functions, allowing them to be represented by a single continuously differentiable control barrier function \cite{RABIEE2025111921}.

In the more specific context of GVFs, \cite{nunes2026safe} employs a smooth approximation of the robot-to-obstacle distance to augment the path-following vector field of \cite{AdrianoIEEE} with obstacle-avoidance capabilities. This regularization is necessary because the Euclidean distance to an obstacle may be non-differentiable, whereas its derivative is required by the safety controller. Nevertheless, the path-following component remains based on the original closest-point construction for the target curve and therefore retains the regularity and numerical implementation issues discussed above. Furthermore, smoothing only a scalar distance function would not, by itself, regularize the selection of the closest point and of its corresponding tangent vector. In contrast, the DEO regularizes the complete closest-point-evaluation operation and can therefore be applied simultaneously to both the normal and tangential components of the GVF in  \cite{AdrianoIEEE}.

Other approaches can also generate sufficiently smooth GVFs, although under different formulations. In \cite{chen2025discrete}, for example, a closed  path represented by discrete data is approximated through a finite Fourier series. The path parameter is then introduced as an additional virtual state $\theta$, and a GVF is constructed in the resulting augmented space. The vector field can be smooth in this augmented state space, but the physical velocity command depends on both the physical configuration and the additional virtual variable. Hence, it is not an autonomous vector field defined solely on the original physical configuration space. The present method, by contrast, constructs an everywhere infinitely differentiable static vector field directly in the original configuration space, without introducing a path-parameter state.

A related probabilistic approach is proposed in \cite{chen2026guiding}, where a score-based diffusion model learns an attracting vector field from discrete path samples, while a second neural network learns a tangential field that induces motion along the path. This formulation accommodates unordered and branching path representations. Our normal component also admits a score interpretation through Gaussian smoothing. Here, however, we consider a prescribed regular curve and its tangent vectors, and compute both components directly through normalized Gaussian-weighted integrals, without training.

We establish the differentiability and limiting properties of the proposed operator and analyze the dynamics generated by the resulting GVF. In particular, for sufficiently small $r$, we show that almost all trajectories converge to a tube around the target curve that shrinks with $r$, and that as $r\rightarrow +0$, we recover the prescribed target curve. We also present a numerical implementation that preserves the smoothness of the construction and experimentally validate the method on a real robotic platform.

In summary, the \textbf{contributions of this paper}  are:

\begin{itemize}
    \item \textbf{Practical}: our approach offers two main benefits. First, the GVF is infinitely differentiable, so cascaded controllers that require its derivatives are not subject to disturbances caused by points of nondifferentiability. Second, implementing the method in \cite{AdrianoIEEE} typically requires sampling points on the target curve. To have a good performance (i.e., guaranteeing convergence and tracking and not having control inputs that are too noisy), a large number of sample points may be necessary, especially when computing derivatives, which can increase the computational cost. Our approach allows the GVF to be implemented using  fewer sampled points at the cost of modifying the target curve, with this trade-off controlled by the parameter $r$. Although the computational cost per sample of our approach is higher than in \cite{AdrianoIEEE}, we show through extensive simulations that this trade-off can be very favorable: the computational time can drop sharply while keeping a very reasonable tracking error. Furthermore, as we will show, we obtain very competitive results even when, instead of implementing \cite{AdrianoIEEE} using discrete points, we use polynomial segments obtained through regression.

    \item \textbf{Theoretical}: for static curves, our GVF generalizes the formulation in \cite{AdrianoIEEE}, which is recovered in the limit $r \to 0^+$. The stability analysis in \cite{AdrianoIEEE} must account for discontinuities in the vector field, which requires a cumbersome mathematical analysis. We establish stability for every $r>0$ without this complication, since the resulting vector field is infinitely differentiable. In essence, our formulation smooths out the discontinuities in \cite{AdrianoIEEE} through an adjustable smoothing parameter, enabling a more straightforward stability analysis. Finally, interestingly, we found that our proof follows more directly if instead of the classical Lyapunov arguments, the \emph{dual} Lyapunov argument of \cite{Rantzer2001} is applied. 
\end{itemize}

\subsection{Basic Definitions}

We use $\mathbb{R}^+$ for the set of nonnegative reals. We explicitly indicate vector dimensions. Thus, instead of writing $v \in \mathbb{R}^{n \times 1}$, we make it clear that it is, for example, a column vector by writing $v \in \mathbb{R}^{n \times 1}$. Given a square matrix $M$, $\text{Tr}(M)$ is its trace and $\text{det}(M)$ its determinant. Given a function $f: \mathbb{R}^{a \times 1} \mapsto \mathbb{R}^{b \times 1}$, we use the Jacobian convention that $\frac{\partial f}{\partial p}: \mathbb{R}^{a \times 1} \mapsto \mathbb{R}^{b \times a}$. We use $M^{\top}$ to denote the transpose of a matrix. $\|\cdot\|$ denotes the Frobenius norm of a matrix/vector. For a variable $y \in \mathbb{R}^n$, a function $F: \mathbb{R}^n \mapsto \mathbb{R}^m$ and a differentiable one dimensional set $\mathcal{Y} \subseteq \mathbb{R}^n$, integrals like $\int_{\mathcal{Y}} F(y)\|dy\|$ should be interpreted as taking any parametrization $y: [0,L] \mapsto \mathbb{R}^n$ of $\mathcal{Y}$, and evaluating $\int_{0}^L F\big(y(\tau)\big)\|y'(\tau)\|d\tau$. Any parametrization will provide the same result.


\section{Methodology}

\subsection{The Original Vector Field}
We begin with some important definitions. We consider a Euclidean configuration space, with $p \in \mathbb{R}^{n \times 1}$ denoting a point in this space represented as a column vector.

\begin{definition}
We introduce the following definitions:
\begin{itemize}
    \item A curve $\mathcal{C} \subset \mathbb{R}^{n \times 1}$ is said to be \emph{regular} if it is bounded, has finite length $\ell_{\mathcal{C}}$, is closed, one-dimensional, at least twice differentiable, has no self-intersections, and has finite maximum curvature $\kappa_{\mathcal{C}}$.
    
    \item We define $p^*_{\mathcal{C}}: \mathbb{R}^{n \times 1} \mapsto 2^{\mathbb{R}^{n \times 1}}$ as the function that maps a point $p$ to the set of all points $\xi \in \mathcal{C}$ that are closest to $p$. Let $\mathcal{U}_{\mathcal{C}} \subset \mathbb{R}^{n \times 1}$ be the set of points $p$ for which $p^*_{\mathcal{C}}(p)$ is a singleton, that is, a set containing a single element (i.e., the complement of the medial axis).
    
    \item Let $T_{\mathcal{C}}: \mathcal{C} \mapsto \mathbb{R}^{n \times 1}$ be the function that maps each point on the regular curve $\mathcal{C}$ to its normalized tangent vector, according to a given predefined orientation.
    
    \item Let $D_{\mathcal{C}}: \mathbb{R}^{n \times 1} \mapsto \mathbb{R}^+$ be the \emph{half-squared} Euclidean distance from $p$ to $\mathcal{C}$:
    \begin{equation}
        D_{\mathcal{C}}(p) \triangleq \min_{\xi \in \mathcal{C}} \frac{\|\xi-p\|^2}{2}.
    \end{equation}
    $\blacksquare$
\end{itemize}
\end{definition}

The GVF $\Psi_{\mathcal{C}}: \mathcal{U}_{\mathcal{C}} \mapsto \mathbb{R}^{n \times 1}$ proposed in \cite{AdrianoIEEE} for tracking a regular curve $\mathcal{C}$, in the special case of static curves and constant gains, can be constructed as follows. For positive scalars $G$ and $H$, it is defined as a weighted sum of \emph{Normal} and \emph{Tangent} components, with the former responsible for convergence to the curve and the latter for traversing it. Precisely,
\begin{equation}
\label{eq:PsiC}
    \Psi_{\mathcal{C}}(p) \triangleq G \underbrace{\big(p_{\mathcal{C}}^*(p)-p\big)}_{\mbox{Normal}} + H \underbrace{T_{\mathcal{C}}\big(p^*_{\mathcal{C}}(p)\big)}_{\mbox{Tangent}}.
\end{equation}

It is shown in \cite{AdrianoIEEE} that, if $\dot{p}(t) = \Psi_{\mathcal{C}}\big(p(t)\big)$, then $D_{\mathcal{C}}(p(t)) \rightarrow 0$ as $t \rightarrow \infty$ for almost all initial conditions $p(0)$, and that the curve $\mathcal{C}$ is traversed according to the orientation chosen for $T_{\mathcal{C}}(\xi)$.

Although this GVF works well in many situations, it has an important practical limitation. Implementing this vector field in a control loop often requires derivatives of $\Psi_{\mathcal{C}}$. For instance, if the control input of the system is acceleration, $\ddot{p}=u$, we can use
\begin{equation}
    u(p,\dot{p}) = K_v\big(\Psi_{\mathcal{C}}(p)-\dot{p}\big) + \frac{d}{dt}\Psi_{\mathcal{C}}(p)
\end{equation}
\noindent for $K_v>0$. The GVF is undefined when $p \notin \mathcal{U}_{\mathcal{C}}$. Although this set has measure zero and this issue may therefore be regarded as a mere technicality, the field $\Psi_{\mathcal{C}}$ can vary rapidly near such points, causing $\frac{d}{dt}\Psi_{\mathcal{C}}(p)$ to become very large. Furthermore, the situation is aggravated by the fact that, when computing this vector field, it is seldom possible to determine the closest point $p^*_{\mathcal{C}}(p)$ analytically; one must instead resort to discretizations and numerical algorithms. These algorithms introduce numerical errors. In this case, when an approximation such as
\begin{equation}
    \frac{d}{dt}\Psi_{\mathcal{C}}(p) \approx \frac{\Psi_{\mathcal{C}}\big(p+\dot{p}\Delta t\big) - \Psi_{\mathcal{C}}\big(p-\dot{p}\Delta t\big)}{2\Delta t}
\end{equation}
\noindent is used, the numerical errors in each computation of $\Psi_{\mathcal{C}}(p)$ can cause disastrous results in the computation of the numerical derivative, because the numerical error may be of an order smaller than $\mathcal{O}(\Delta t)$.

At the cost of not tracking the desired curve $\mathcal{C}$ perfectly, we propose a modification of $\Psi_{\mathcal{C}}(p)$ that guarantees that it is defined and infinitely differentiable everywhere, while also behaving well under numerical differentiation. This trade-off between tracking accuracy and smoothness will be governed by a parameter $r$.

\subsection{The Modified Vector Field}

We begin by noting that the vector field in \cite{AdrianoIEEE} can be constructed by defining a specific function $\psi_{\mathcal{C}}: \mathcal{C} \times \mathbb{R}^{n \times 1} \mapsto \mathbb{R}^{n \times 1}$ as
\begin{equation}
    \label{eq:psic}
    \psi_{\mathcal{C}}(\xi,p) \triangleq G(\xi-p)+HT_{\mathcal{C}}(\xi)
\end{equation}
for positive constants $G,H$, and \emph{evaluating} $p^*_{\mathcal{C}}(p)$ into the variable $\xi$. The source of nondifferentiability is that $p^*_{\mathcal{C}}(p)$ is not uniquely defined at points $p \notin \mathcal{U}_{\mathcal{C}}$. Furthermore, it varies rapidly near such points. We proceed by replacing this \emph{closest-point-evaluation} operation with one that is better behaved, that is, differentiable. For this purpose, we need some key definitions.

\begin{definition}\label{def:deo}
    Consider a regular curve $\mathcal{C}$ and let $r$ be a positive number. Define the function $\rho_{\mathcal{C}}^{r}: \mathcal{C} \times \mathbb{R}^{n \times 1} \mapsto \mathbb{R}^+$ as
    \begin{equation}
        \rho_{\mathcal{C}}^{r}(\xi,p) \triangleq \frac{\exp\left(-\frac{\|\xi-p\|^{2}}{2r^2}\right)}{\int_{\mathcal{C}} \exp\left(-\frac{\|\chi-p\|^2}{2r^2}\right)\|d\chi\|}.
    \end{equation}
    
    We then define the $r$-\emph{Differentiable Evaluation Operator (DEO)}, which maps a function $f: \mathcal{C} \times \mathbb{R}^{n \times 1} \mapsto \mathbb{R}^{a \times b}$ to a function $\overline{f}_r: \mathbb{R}^{n \times 1} \mapsto \mathbb{R}^{a \times b}$ defined by $\overline{f}_r(p)=\Ee\big[f(\xi,p)\big](p)$, where
    \begin{equation}
       \Ee\big[f(\xi,p)\big](p) \triangleq \int_{\mathcal{C}} f(\xi,p)\rho_{\mathcal{C}}^{r}(\xi,p)\|d\xi\|.
    \end{equation}
    Sometimes, we omit the argument ``$(p)$'' and simply write $\Ee\big[f(\xi,p)\big]$.
    $\blacksquare$
\end{definition}

The DEO is linear and approximately implements the operation $f(\xi,p) \mapsto f\big(p^*_{\mathcal{C}}(p),\,p\big)$, that is, the closest-point-evaluation operation in $\xi$, while preserving $k$-times differentiability with respect to $p$, provided that $f(\xi,p)$ itself is $k$-times differentiable with respect to $p$. In particular, if $f$ is infinitely differentiable with respect to $p$, then so is $\Ee[f(\xi,p)](p)$. Under some conditions on $f$ and, at least, for $p \in \mathcal{U}_{\mathcal{C}}$, the DEO reduces to the closest-point-evaluation operation as $r \rightarrow +0$; see Lemma \ref{lemma:convEe} in the Appendix.

The choice of the symbol $\mathbb{E}$, which resembles the expectation operator for a random variable, is not accidental. The quantity $\Ee\big[f(\xi,p)\big](p)$ can be interpreted as the expectation of the function $f(\Xi,p)$ with respect to a $p$-parametrized random variable $\Xi(p) \in \mathcal{C}$ distributed according to the $p$-parametrized distribution $\Xi(p) \sim \rho_{\mathcal{C}}^{r}(\xi,p)$. As $r \rightarrow +0$, $\rho_{\mathcal{C}}^{r}(\xi,p)$ converges to a Dirac delta distribution concentrated at $p_{\mathcal{C}}^*(p)$, namely, $\delta(\xi-p_{\mathcal{C}}^*(p))$. Equivalently, the random variable becomes ``deterministic,'' with $\Xi(p)=p_{\mathcal{C}}^*(p)$. This interpretation provides useful intuition about the operator.

Using this construction, we can define an \emph{infinitely differentiable version of} \eqref{eq:PsiC}, namely, the function $\Psi_{\mathcal{C}}^{r}: \mathbb{R}^{n \times 1} \mapsto \mathbb{R}^{n \times 1}$ given by
\begin{equation}
    \label{eq:PsiCr}
    \Psi_{\mathcal{C}}^{r}(p) \triangleq \Ee\big[\psi_{\mathcal{C}}(\xi,p)\big](p).
\end{equation}
In Subsection \ref{subs:implementation}, we show how this quantity can be approximated with arbitrary accuracy. We now state the main result of this paper.

\begin{proposition} \label{prop:mainprop}
    We have the following properties:
    \begin{itemize}
        \item For $r>0$, $\Psi_{\mathcal{C}}^{r}(p)$ is infinitely differentiable with respect to $p$ everywhere.
        
        \item For all $p \in \mathcal{U}_{\mathcal{C}}$,
        \begin{equation}
            \lim_{r \rightarrow +0} \Psi_{\mathcal{C}}^{r}(p) = \Psi_{\mathcal{C}}(p).
        \end{equation}
        
        \item There exists an $r_{\max} > 0 $ such that, for $r \leq r_{\max}$ and for almost all initial conditions, that is, except for a set of measure zero, the dynamical system $\dot{p}=\Psi_{\mathcal{C}}^{r}(p)$, $p \in \mathbb{R}^{n \times 1}$ enters in finite time a neighborhood $\mathcal{C}^r \supseteq \mathcal{C}$ of $\mathcal{C}$ and stays there forever,  moving inside it, never stopping. Furthermore, $\lim_{r \rightarrow +0} \mathcal{C}^r = \mathcal{C}$.

    \end{itemize}
\end{proposition}
\begin{proof}
    See the Appendix. $\blacksquare$
\end{proof}

Proposition 1 says that the vector field is infinitely differentiable, that it approaches the vector field in \cite{AdrianoIEEE} in the limit $r \rightarrow +0$ and that it is guaranteed to drive the system, for almost all initial conditions, to a neighborhood $\mathcal{C}^r$ of $\mathcal{C}$, that can be shrunk to be arbitrarily close to $\mathcal{C}$.

This vector field provides a trade-off between perfectly tracking the desired curve $\mathcal{C}$ and obtaining smoother, and in fact infinitely differentiable, control inputs. As $r$ increases, the tracking error relative to $\mathcal{C}$ increases, whereas the derivatives of $\Psi_{\mathcal{C}}^{r}(p)$ with respect to $p$ decrease in magnitude.

Finally, it is important to make the following comparison with the result in \cite{AdrianoIEEE}: our proposed approach reduces to  \cite{AdrianoIEEE} for static curves and constant gains. The case of time varying curves will be discussed in  future work. Considering the gains, the full form of the vector field in \cite{AdrianoIEEE} for static curves is:
\begin{equation}
\label{eq:PsiC_full}
    \Psi_{\mathcal{C}}(p) = G(p) \big(p_{\mathcal{C}}^*(p)-p\big)+H(p)T_{\mathcal{C}}\big(p^*_{\mathcal{C}}(p)\big).
\end{equation}
In comparison with \eqref{eq:PsiC}, note that $G(p)$ and $H(p)$ are allowed to be nonnegative functions of $p$, whereas in our case $G$ and $H$ are constants. This flexibility, not present in our work, can be helpful to tune the controller. We believe our theoretical guarantee can be extended to also have functions and this will be investigated in a future work, nevertheless, note that although we cannot fully modify $G(p)$ and $H(p)$, we can trivially extend our result to have more flexibility  than simply having a constant $G$ and $H$. This is because if the whole vector field is multiplied by a $k$-times positive differentiable function $v: \mathbb{R}^n \mapsto \mathbb{R}^+$ the resulting vector field $\hat{\Psi}_{\mathcal{C}}^r(p) = v(p)\Psi_{\mathcal{C}}^r(p)$ is still $k$-times differentiable and keep the convergence properties. The latter fact comes because a multiplication by $v(p)>0$ just parametrizes the speed at which the robot will be moving,  and thus do not modify the theoretical convergence guarantees.

As discussed in \cite{AdrianoIEEE}, in some applications, e.g., fixed wing aircrafts, it is desirable to have constant speed. We could achieve this by selecting $v(p) = v_{c}/\|\Psi_{\mathcal{C}}^r(p)\|$. But then the vector field no longer can be guaranteed to be even continuous, because there can be points $p$ in which $\Psi_{\mathcal{C}}^r(p)=0$. This is specially the case with closed curves, and is not a limitation of our work, but a topological constraint. There is no continuous nonvanishing autonomous (i.e., non time-varying) vector field in $\mathbb{R}^n$ whose globally asymptotically stable limit cycle is a closed curve \cite{yao2022topological}.

\subsection{Numerical Implementation}
\label{subs:implementation}

Analytical computation of the integral in $\Psi_{\mathcal{C}}^{r}(p)$ is usually difficult. However, it can be approximated as follows. Sample $N$ evenly spaced points $\xi_i$ on the curve, such that $\|\xi_{i+1}-\xi_i\|$ is approximately constant. Then, for a function $f(\xi,p)$,
\begin{equation}
    \Ee[f(\xi,p)](p) \approx \sum_{i=1}^N \tilde{\rho}_{\mathcal{C}}^r(i,p)f(\xi_i,p)
\end{equation}
\noindent where
\begin{equation}
    \tilde{\rho}_{\mathcal{C}}^r(i,p) \triangleq \frac{\exp\left(-\frac{\|p-\xi_i\|^{2}}{2r^2}\right)}{\sum_{j=1}^N \exp\left(-\frac{\|p-\xi_j\|^{2}}{2r^2} \right)}.
\end{equation}
To avoid numerical instabilities that may occur when all the terms $\exp\left(-\frac{\|p-\xi_i\|^{2}}{2r^2}\right)$ are too small, it is preferable to compute
$\tilde{D}_{\mathcal{C}}(p) = \min_{j=1,\ldots,N} \|p-\xi_j\|^2$
and instead use the mathematically equivalent expression
\begin{equation}
    \tilde{\rho}_{\mathcal{C}}^r(i,p) \triangleq \frac{\exp\left(\frac{\tilde{D}_{\mathcal{C}}(p)-\|p-\xi_i\|^{2}}{2r^2}\right)}{\sum_{j=1}^N \exp\left(\frac{\tilde{D}_{\mathcal{C}}(p)-\|p-\xi_j\|^{2}}{2r^2} \right)}.
\end{equation}
This approximation is analogous to computing the integrals using the rectangle rule. More accurate results could be obtained using more sophisticated integration rules. However, it is important to keep the points $\xi_i$ fixed. This ensures that the approximation $\tilde{\Psi}_{\mathcal{C}}^r(p)$ of $\Psi_{\mathcal{C}}^{r}(p)$ remains infinitely differentiable with respect to $p$.
\section{Experimental Results}
\subsection{The Model}
\label{subs:model}
Throughout this section, we will perform experiments and simulations to evaluate the claimed benefits of our proposed method, and also to help the user to select these parameters. 

\begin{figure}[h!]
    \centering
    \includegraphics[width=0.45\linewidth]{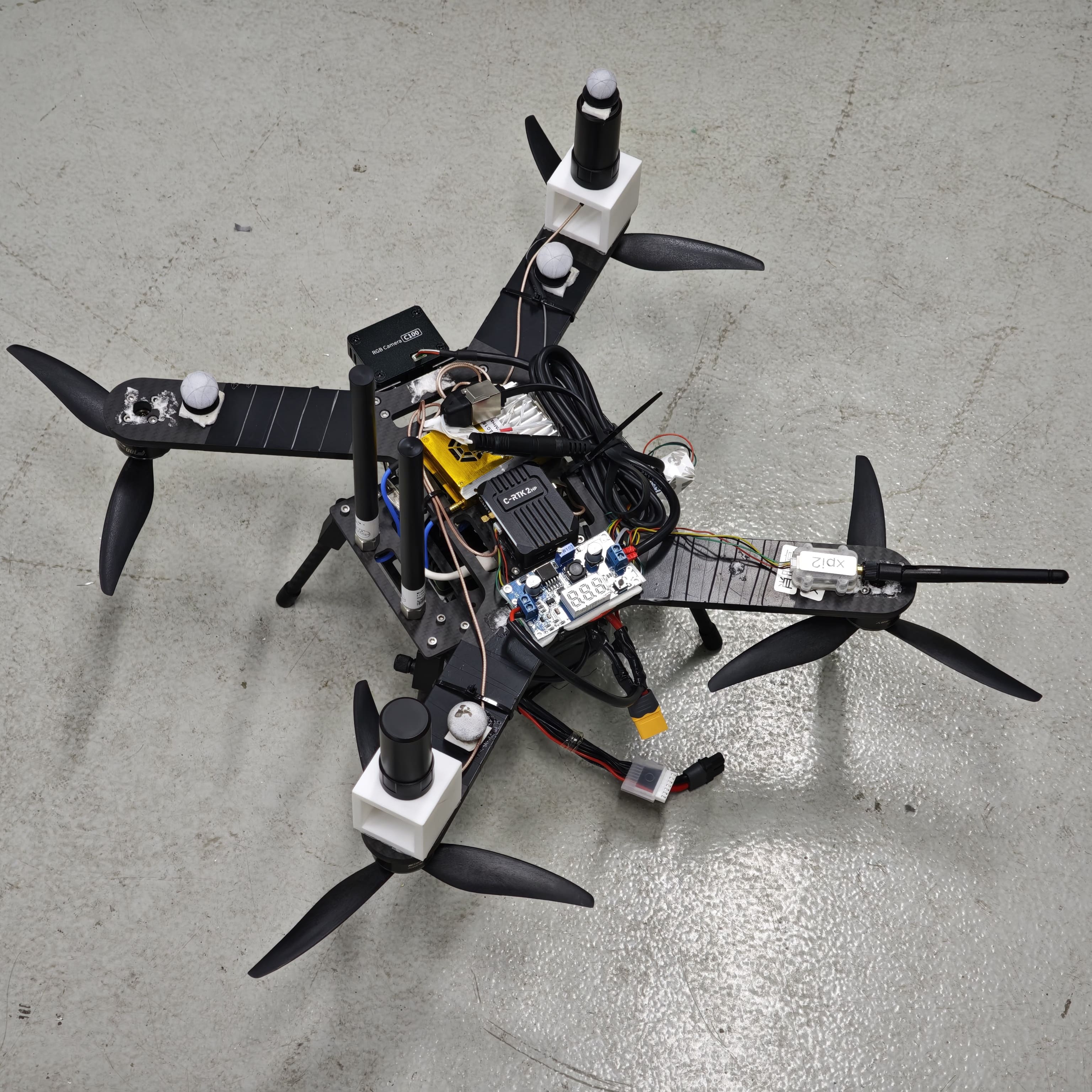}
        \includegraphics[width=0.45\linewidth]{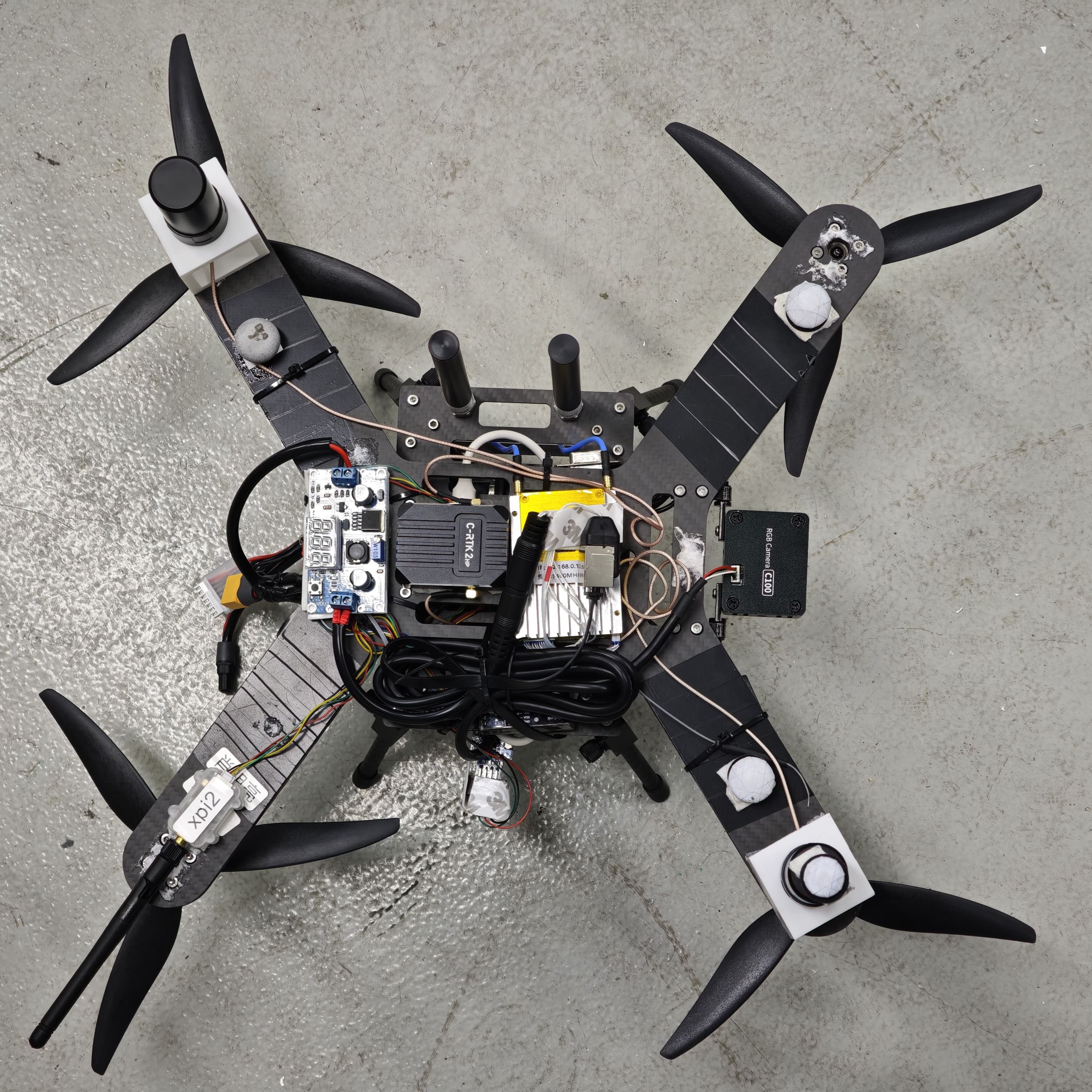}
    \caption{Drone used in the experiment and simulation}
    \label{fig:drone}
\end{figure}

We use a custom-built quadrotor platform (see Figure \ref{fig:drone}). It uses a low-level controller to handle linear velocity commands, transforming them into angular velocity and total thrust inputs. We will present the model of the drone plus the low level controller in this subsection, as well as the parameters. The model and parameters used in simulation and real experiments are the same.

We consider a drone that is controlled through \emph{total thrust} $T \in \mathbb{R}$ and  \emph{angular velocity vector} $\omega \in \mathbb{R}^{3 \times 1}$, with the latter written in inertial frame. Let $p, v, z \in \mathbb{R}^{3 \times 1}$ be, respectively, the position, linear velocity and $z$ axis (aligned with the total thrust direction) of the drone measured  the inertial frame. We also consider a virtual state $v_d \in \mathbb{R}^{3 \times 1}$ needed to implement our controller. We assume the state evolving equations:
\begin{equation}
\label{eq:dynsys}
    \dot{p} = v \ , \ \dot{v} = g + \frac{s(T)}{M}z \ , \ \dot{v}_{d} = \nu \ , \ \dot{z} = \omega \times z 
\end{equation}

\noindent in which $g$ is the gravity vector, measured in the inertial frame, $M$ is the drone mass and $\nu$ is a virtual control variable. The function $s(T)$ implements a saturation of $T$ between $T_{\text{min}}$ and $T_{\text{max}}$.

Following \cite{GONCALVES2024104601}, given a  velocity vector field $\Psi_r(p)$ such that $v = \Psi_{\mathcal{C}}^r(p)$ is desired, we use the following control scheme, for positive constants $K_{v}, K_{\Delta}$ and $K_{\omega}$:

\begin{eqnarray}
\label{eq:contsys}
 && a_{d}(p,v) \triangleq K_v\big(\Psi_{\mathcal{C}}^r(p)-v\big) + \frac{d}{dt} \Psi_{\mathcal{C}}^r(p) \nonumber \\
  && a_{c}(p,v,v_d) \triangleq a_{d} {-} g {+} K_{\Delta}\big(v_d{-}v\big) \ , \  z_{d}(p,v,v_d) \triangleq \frac{a_{c}}{\|a_c\|}  \nonumber \\
 && \nu(p,v,v_d) = a_{d}  \ , \  T (p,v,z,v_d)= M z^{\top} a_{c} \nonumber \\
 && \omega(p,v,z,v_d) = z_{d} \times \Big(\frac{d}{dt} z_d - K_{\omega}z\Big).
\end{eqnarray}
 Note that this control scheme requires the \emph{second} derivative of the field $\Psi_{C}^r(p)$. 

Even though the terms $\frac{d}{dt} \Psi_{C}^r$ and $\frac{d}{dt}z_d$ could be computed analytically, we  estimated them numerically for the sake of simplicity and computational efficiency, also aiming to test the claim that our implementation is suitable for numerical differentiation. They are implemented as follows:

\begin{eqnarray}
&& \frac{d}{dt} \Psi_{C}^r(p) \approx \frac{\Psi_{C}^r(p+v\Delta t)-    \Psi_{C}^r(p-v\Delta t)}{2\Delta t} \nonumber \\
&& z_d^+ = z_d(p+v\Delta t, v+a\Delta t, v_d+a_d\Delta t)\nonumber \\
&& z_d^- = z_d(p-v\Delta t, v-a\Delta t, v_d-a_d\Delta t) \nonumber \\
&& \frac{d}{dt} z_d(p,v,v_d) \approx \frac{z_d^+-z_d^-    }{2\Delta t}
\end{eqnarray}

\noindent for a small $\Delta t$. In this case, we use the drone acceleration given by $a = \dot{v} = g + (s(T)/M)z$.

The parameters used (real experiment and simulation) can be seen in Table \ref{tab:parameters}. The target curve $\mathcal{C}$, the value of $r$ and the number of sampled points $N$ will change during the simulations and experiments and thus are not reported in this table.




\begin{table}[ht]
    \centering
    \caption{Parameter values (simulation and experiment).}
    \label{tab:parameters}
    \begin{tabular}{cc|cc|cc}
        \hline
        Name & Value & Name & Value & Name & Value \\
        \hline
        $G$ & $2.0\,\mathrm{s}^{-1}$
        & $H$ & $1.0\,\mathrm{m/s}$
        & $K_v$ & $2.0\,\mathrm{s}^{-1}$ \\

        $K_{\Delta}$ & $2.0\,\mathrm{s}^{-1}$
        & $K_{\omega}$ & $2.0\,\mathrm{s}^{-1}$
        & $M$ & $2.09\,\mathrm{kg}$ \\

        $\Delta t$ & $0.001s$
        & $T_{\min}$ & $0\,\mathrm{N}$
        & $T_{\max}$ & $30\,\mathrm{N}$ \\
        \hline
    \end{tabular}
\end{table}

\subsection{Choosing the Parameters of the GVF}
\label{subs:analysis}

The main design choice considered in this paper is the smoothing parameter $r$, including whether to use $r=0$, corresponding to the original approach in~\cite{AdrianoIEEE}, or $r>0$, corresponding to our proposed approach. Additional choices arise in the numerical implementation. For $r>0$, we use the sampled-point approximation described in Subsection~\ref{subs:implementation} and must select the number of samples $N$. For $r=0$, alternatives to searching through sampled points are available. Here, we also consider \emph{piecewise polynomial approximation}: we use regression to approximate $\mathcal{C}$ with $N$ polynomial segments of degree $m$, matching derivatives through order $m-1$ at the joins, and search for the closest point on each segment. For a unified notation, we use $m=0$ to denote the sampled-point representation and $m>0$ to denote polynomial segments. For $m\geq1$, finding the closest point on a polynomial segment requires finding the real roots of a polynomial of degree at most $2m-1$. Numerical root-finding algorithms can be used. For $m=1$ and $m=2$, the stationary-point equations are linear and cubic, respectively, and can be solved analytically, using Cardano's formula in the cubic case.

These choices depend on several factors, including the target curve, the GVF parameters, and the low-level dynamics. We must balance steady-state tracking error, measured relative to the original curve, against computational time. In this subsection, we present a case study using the model in Subsection~\ref{subs:model}, providing practical observations and simulation results to help guide these choices.

We first define the performance metrics. Let $\Phi(t,p_0,\mathcal{C},\Pi)$ denote the position $p(t)$ obtained by simulating the system in~\eqref{eq:dynsys} and~\eqref{eq:contsys}, with $p(0)=p_0$, $z(0)$ pointing upwards, and the remaining state variables initially zero. The controller parameters are those in Table~\ref{tab:parameters}, the target curve is $\mathcal{C}$, and the implementation parameters are collected in $\Pi=(r,N,m)$. This bundle contains the smoothing parameter $r\geq0$, the number $N$ of points or polynomial segments, and the polynomial degree $m$, with $m=0$ denoting points. For simplicity, we refer to both points and polynomial segments as ``segments'' when discussing their computational cost. We only consider $m=0$ when $r>0$, since the implementation studied here uses sampled points.\footnote{A piecewise polynomial representation could also be used for the smoothed approach. Evaluating the resulting integrals would generally require additional numerical quadrature or special functions. We leave the assessment of such implementations for future work.}

We define the \emph{average steady-state tracking error} as \footnote{Since $D_{\mathcal{C}}$ is the half-squared distance, $\sqrt{2D_{\mathcal{C}}}$ is the Euclidean distance to the curve.}
\begin{equation}
\label{eq:E_def}
    E(\Pi) \triangleq \limsup_{T\rightarrow\infty}\frac{1}{T}\int_0^T
    \mathbb{E}\left[\sqrt{2D_{\mathcal{C}}\big(\Phi(t,P_0,\mathcal{C},\Pi)\big)}\right]dt.
\end{equation}
Here, the expectation is taken over the initial position $P_0$ and the target curve $\mathcal{C}$, conditioned on successful tracking for the given $\Pi$. Their distributions will be specified soon. A trajectory is classified as successful if its steady-state tracking error is below $20\,\mathrm{cm}$ and it traverses the curve rather than becoming stuck. Thus, $E(\Pi)=E(r,N,m)$ measures the average steady-state error among successful trajectories. We report the failure rate separately as $F(\Pi)\in[0,1]$.  Throughout this analysis, this distance is measured relative to the \emph{original target curve} $\mathcal{C}$, evaluated numerically using a dense reference sampling, rather than relative to the polynomial approximation or the deformed set $\mathcal{C}^r$ caused by smoothing (see Proposition \ref{prop:mainprop}).

First, consider the ideal case without discretization:

\emph{\textbf{Practical Observation 1}: Even without discretization, small values of $r$ can produce large derivatives of the guidance signal, particularly near the medial axis. In a controller that requires these derivatives, actuator limits, finite bandwidth, and numerical approximations can then degrade tracking or cause failures. On the other hand, large values of $r$ can introduce excessive smoothing and thus cause large errors as well. The choice of $r$ must balance these effects.}

The difficulties associated with small $r$ depend on the low-level dynamics, including the number of GVF derivatives required and the system's sensitivity to them. Consequently, a positive value of $r$ that performs well for one platform need not be suitable for another. For sufficiently small $r$, Lemma~\ref{lemma:persist} bounds the size of an invariant neighborhood of $\mathcal{C}$ for the ideal first-order dynamics. However, this bound depends strongly on the curve geometry and does not directly predict the tracking error of the full closed-loop system.

Dimensional analysis of the ideal GVF provides another useful starting point. For geometrically similar curves, scaling all spatial coordinates by a factor $a$ and scaling both $r$ and $H/G$ by the same factor preserves the relative geometry of the ideal first-order trajectories, with correspondingly scaled initial conditions. For example, if a particular choice produces an error equal to $5\%$ of the curve length, doubling the curve size suggests doubling $r$ and $H/G$ to retain the same relative behavior. Thus:

\emph{\textbf{Practical Observation 2}: The value of $r$ should be commensurate with the ``size'' of the curve. Larger curves require larger values of $r$.}

With discretization, computational time must also be balanced against performance. Small values of $N$, or low polynomial degrees when $r=0$, reduce evaluation time but may produce poor tracking or failures, such as becoming stuck. Larger values can improve the approximation, at the cost of additional computation. For the smoothed field, the interaction between $r$ and the sample spacing is particularly important:

\emph{\textbf{Practical Observation 3}: The smoothing parameter $r$ and the number of samples $N$ should be chosen together. For approximately uniform arc-length sampling and being $\ell_{\mathcal{C}}$ the length of the curve, the sample spacing is $h\approx\ell_{\mathcal{C}}/N$. Near the curve, a useful heuristic is to make this spacing small compared with $r$, so that several nearby samples contribute appreciably to the Gaussian-weighted average. If the sampling is too sparse relative to $r$, the weights can become concentrated on individual samples and change rapidly as the robot moves, even though the implemented field remains infinitely differentiable. Smaller values of $r$ therefore generally require denser sampling. The required sampling density depends on the curve geometry and the derivatives required by the controller.}

\begin{figure}[htb]
    \centering
    \includegraphics[width=1\linewidth]{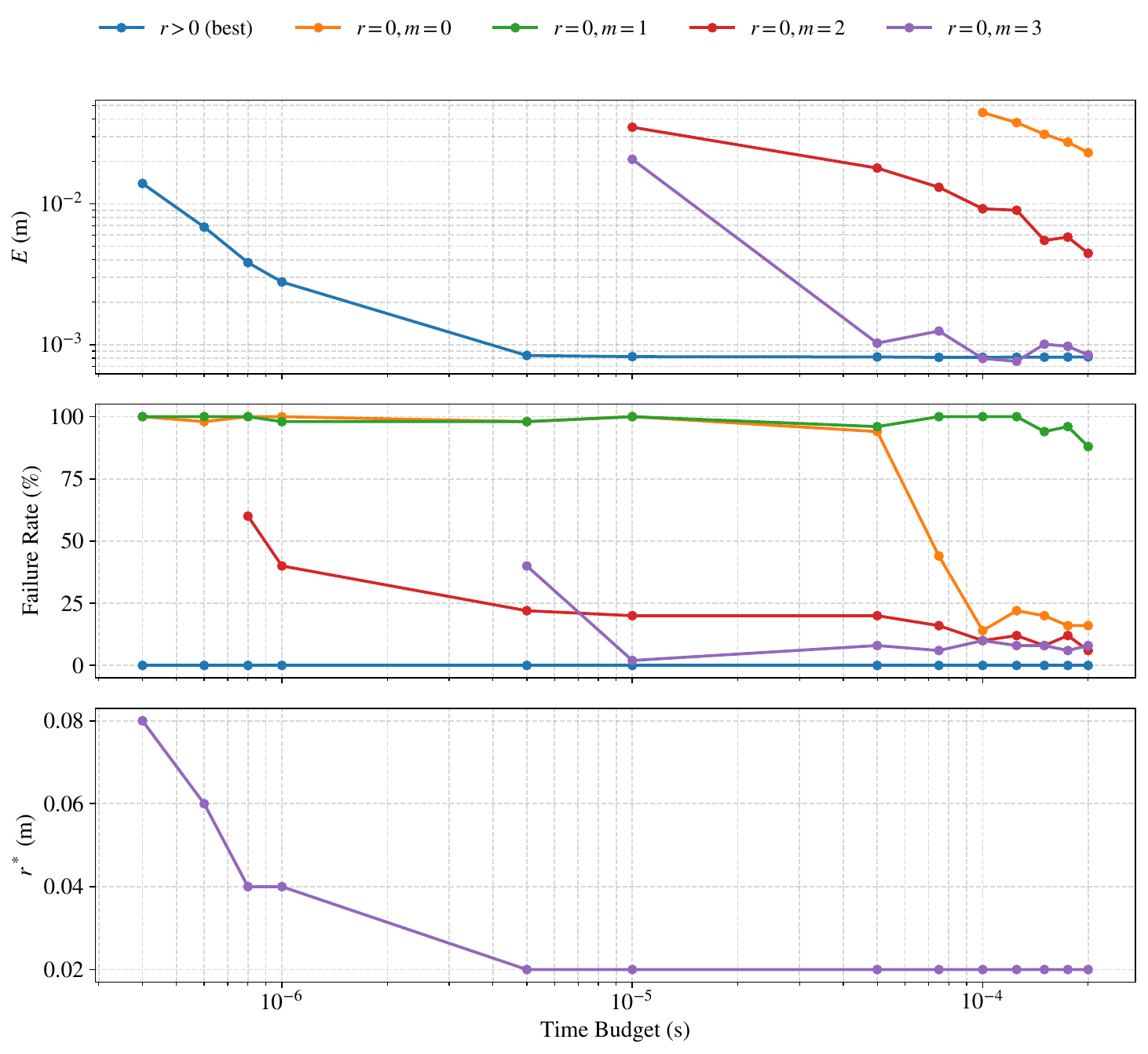}
    \caption{Computational-budget experiment using five target curves and ten initial positions per curve. Top: average steady-state tracking error among successful trajectories. Middle: failure rate. Bottom: best-performing tested smoothing parameter $r>0$. Each implementation uses the largest number of segments estimated to fit within the evaluation budget.}
    \label{fig:budget_experiment}
\end{figure}

For a fixed polynomial degree, all implementations considered here have evaluation complexity $\mathcal{O}(N)$: the smoothed field sums contributions from all sampled points, while the original field searches through all points or polynomial segments.\footnote{Spatial search structures as KD-Tress and BVHs can accelerate nearest-point queries after preprocessing. Here, we compare implementations that evaluate the given representations by direct summation or exhaustive search, without such structures.} Their costs per segment, however, differ substantially. Quantifying these differences helps guide the choice of $\Pi=(r,N,m)$.

\emph{\textbf{Practical Observation 4}: Let $\tau(r,m)$ denote the evaluation cost per segment in our optimized C++ implementations when using a smoothing factor $r$ and polynomials of degree $m$ (for $r=0$). The measured cost ratios are approximately $\tau(r,0)/\tau(0,0)\approx10$ for $r>0$, $\tau(0,1)/\tau(0,0)\approx20$, $\tau(0,2)/\tau(0,0)\approx200$, and $\tau(0,3)/\tau(0,0)\approx600$.}

This means that, for example, implementing our proposed strategy with $N=200$ points is comparable in time to implementing \cite{AdrianoIEEE} using $N = 10 \times 200 = 2,000$ sampled points or with polynomial segments of degree $2$ with $N = 2,000/200 = 10$ polynomial segments. These ratios are empirical estimates obtained from our implementation. They depend on the hardware and numerical algorithms used but they, nevertheless, provide a conservative bound according to our experiments.

We performed simulations using the model in Subsection~\ref{subs:model} and the parameters in Table~\ref{tab:parameters}. The initial position was drawn from a zero-mean normal distribution with covariance $\sigma^2I$, where $\sigma^2=4\,\mathrm{m}^2$. The curve was selected with equal probability from five curves of roughly the same size, centered at the origin and contained in a ball of radius $1\,\mathrm{m}$. We used ten initial positions for each curve, giving 50 trajectories for each parameter bundle $\Pi$, simulated from $t=0$ to $t=30\,\mathrm{s}$. Computations were performed on an 11th Gen Intel Core i7-11800H CPU at 2.30 GHz, with each GVF evaluation using a single core.

We then considered a \emph{computational-budget experiment}: given a time budget $t_b$ for one GVF evaluation on a single CPU core of our machine, which of the tested choices of $r$, $N$, and $m$ provides the best tracking performance? Note that the controller in~\eqref{eq:contsys} requires \emph{nine such} evaluations. Figure~\ref{fig:budget_experiment} presents the results for budgets ranging from $0.1\,\mu\mathrm{s}$ to $200\,\mu\mathrm{s}$. For each feasible choice of $r$ and $m$, we use the largest affordable number of segments, estimated as $N = \lfloor t_b/\tau(r,m)\rfloor$ (see ``Practical Observation 4'' for the meaning of $\tau$). The top panel reports the average error $E$, the middle panel reports the failure $F$, and the bottom panel shows the best-performing tested value, named $r^*$, of $r>0$ at each budget. Note that we only report averages of $E$ when the failure rate $F$ is less than $20\%$ as otherwise the mean would be contaminated.

Figure~\ref{fig:budget_experiment} shows the expected behavior that, the larger is the time budget, the better is the results we can obtain, in the sense of having a smaller average error and failure rate. It also shows a substantial advantage for the smoothed approach in this case study. It achieves reasonable steady-state errors at small computational budgets, while the other implementations require substantially larger budgets to become viable. By that point, the smoothed approach has already reached $E=0.0008\,\mathrm{m}$, an error likely to be below the tracking accuracy achievable on the physical platform described in Subsection~\ref{subs:model}. Furthermore, no failures were observed for the selected smoothed configurations at any tested budget, whereas the other implementations exhibited nonzero failure rates. These results show that, for the tested curves and controller, smoothing can provide good tracking with a simple implementation and a small computational budget.

At large time budgets, the error of the smoothed field levels off at approximately $E=0.0008\,\mathrm{m}$, with a best tested value of $r=0.02\,\mathrm{m}$. For time budgets $t_b > 200\mu s$, the best error for $r>0$ becomes stable at this value, whereas the polynomial approach can potentially obtain  smaller error values such as $0.0001\,\mathrm{m}$. However, these results are unlikely to be useful in practice, since such an error is very unlikely to be obtained in a practical situation in the  quadrotor platform considered here.

\subsection{Smoothness of Numerical Derivatives}

We have claimed and proved that the vector field $\Psi_{\mathcal{C}}^r$ is smooth, and that, compared to the original $\Psi_{\mathcal{C}}$, it behaves much better when computing derivatives. In this subsection, we will illustrate this.

To do so, we assumed a time budget that would allow us to compute \cite{AdrianoIEEE} using $15,000$ sampled points. Per ``Practical Observation 4'' in the previous subsection, this would allow us to use, still with $r=0$, $750$ polynomial segments of degree $1$, $75$ polynomial segments of degree 2 or $25$ polynomial segments of degree $3$. Furthermore, it would allow us to have $1,500$ points for $r>0$.

\begin{figure*}[t!]
    \centering
    \includegraphics[width=\linewidth]{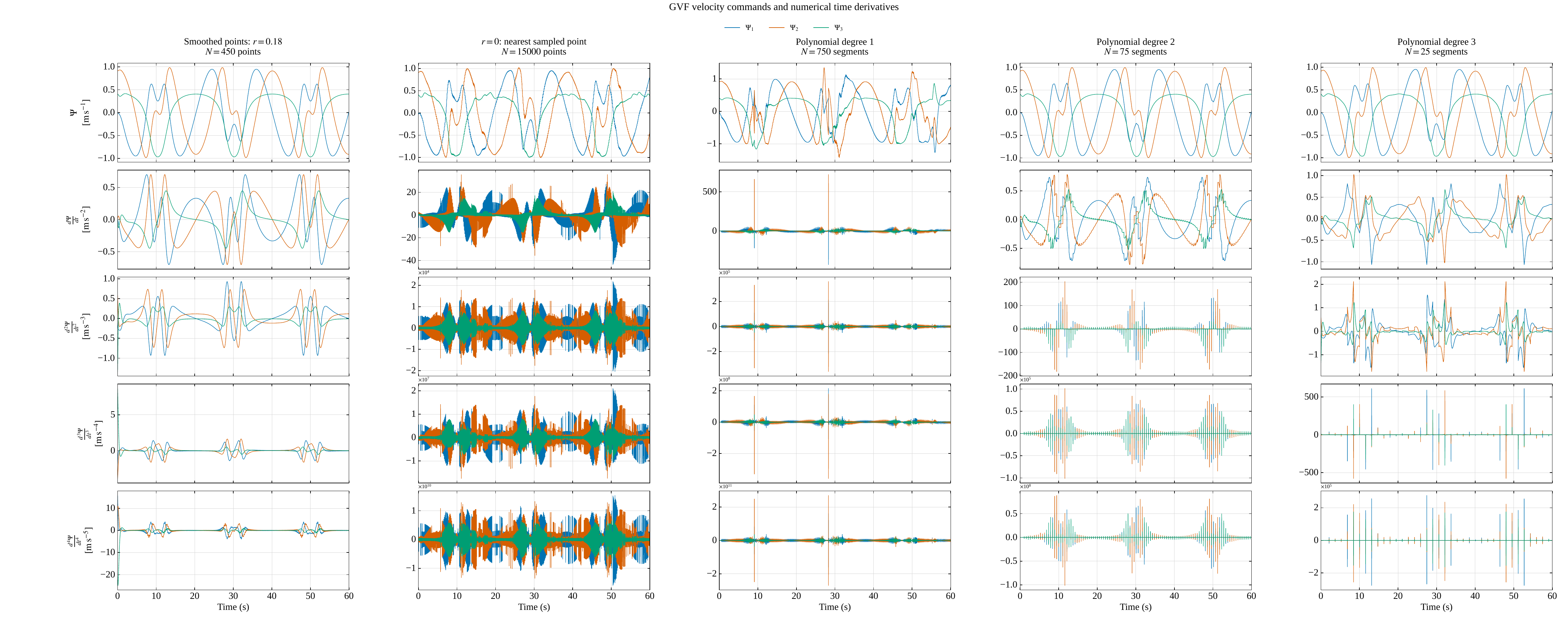}
    \caption{Comparison of the control inputs from the GVF
    and their numerical derivatives considering different
    implementations.}
    \label{fig:comparison2}
\end{figure*}

We have implemented the model in Subsection \ref{subs:model} using these five strategies, and computed the derivatives of the control input $\Psi_{\mathcal{C}}^r$ numerically up  to the $4^{th}$ derivative. In the case of $r>0$, we took $r=0.18$ (as in the upcoming experiment) and, instead of $1,500$ points, we took roughly one third, ($30\%$) which $450$ points, thus \emph{providing one third of the time budget of the other approaches}.

Figure \ref{fig:comparison2} shows the plots of the derivatives up to order $4$ for each one of the approaches, computed with $\Delta t = 0.001s$. The smoothness of our approach is very clear: all high order derivative plots stay ``stable'', whereas in the others the numerical derivatives start to degenerate due to numerical errors that come from the nondifferentiability. For example, when using $r=0$ and sampled points, the control input is barely differentiable even once, having already high frequency noise.

\subsection{Experiment with Drone}

 We implemented the vector field on a custom-built quadrotor in an indoor motion-capture arena measuring approximately 13.8 m × 30.0 m.  The overhead motion-capture system consisted of 54 Swift120A/AW cameras arranged in a regular rectangular layout. Each camera provides 12 MP resolution and operates at up to 320 fps, with a specified positioning accuracy of ±0.03 mm. The FZMotion system was operated using FZMotion's Moca software, which was configured to publish pose data at 120 Hz. The pose data were forwarded through VRPN to the onboard ROS 2/PX4 control stack, with Micro XRCE-DDS providing the ROS 2-PX4 transport. The drone computes everything in a Raspberry Pi 4B with 2 GB RAM. The system model and controller parameters are as described in Subsection \ref{subs:model} and Table \ref{tab:parameters}, respectively.

\begin{figure*}[t]
    \centering
    \includegraphics[trim=0 0 0 1.5cm, clip, width=1\linewidth]{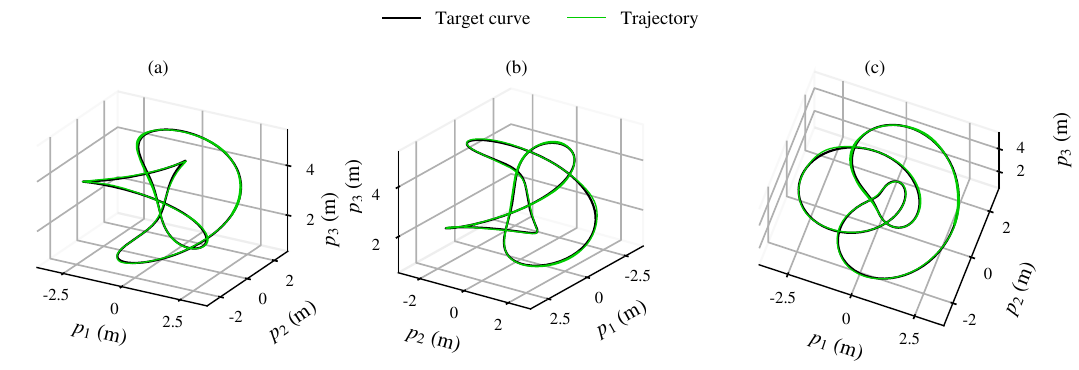}
    \caption{Target path (black) and tracking trajectory (green) in the flight experiment}
    \label{fig:target_path}
\end{figure*}

 Our target curve is shown in Figure \ref{fig:target_path}. It is important to mention that this curve \emph{did not appear as one of the five curves used in the analysis in Subsection  \ref{subs:analysis}}. Furthermore, it is significantly more complex topology-wise and also larger, being contained in a radius of $3 \text{m}$ instead of $1 \text{m}$. These are interesting, because we can then try to extrapolate from the analysis in Subsection  \ref{subs:analysis}.

 The experiments in Subsection  \ref{subs:analysis} were performed on a  11th Gen Intel Core i7-11800H CPU at 2.30 GHz using a single core, whereas the drone has a Raspberry Pi 4B with 2 GB RAM. We estimate that the latter is roughly $8$ times slower than the former. This implies that, if we have to adapt the $x$ axis of Figure \ref{fig:budget_experiment} to the drone, also considering that each evaluation of the full controller requires $9$ evaluations of the GVF, we arrive to the consideration that the time scale in  Figure \ref{fig:budget_experiment} should be multiplied by $100$. If we want to operate at $1 \text{KHz}$, this implies looking at $t_b = 1 \text{ms}/100 = 10 \mu \text{s}$ in Figure \ref{fig:budget_experiment}. This suggests $r=0.02 \text{m}$ and that we would have an error $0.8 \text{cm}$. But our curve is roughly three times the size of the curves considered in Subsection  \ref{subs:analysis}. ``Practical Observation 2'' implies that we should roughly triple $r=0.02 \text{m}$ then and expect roughly  three times the average error. In the real experiment, considering real life behaviors not modeled by the model in Subsection \ref{subs:model}, we decided  to have $r=0.18 \text{m}$. The time budget of $1 \text{ms}$ in the drone allow us to use roughly $N=800$ points. We decided to use $N=300$.

 \begin{figure}[h!]
    \centering
    \includegraphics[trim=0 0 0 0, clip, width=0.9\linewidth]{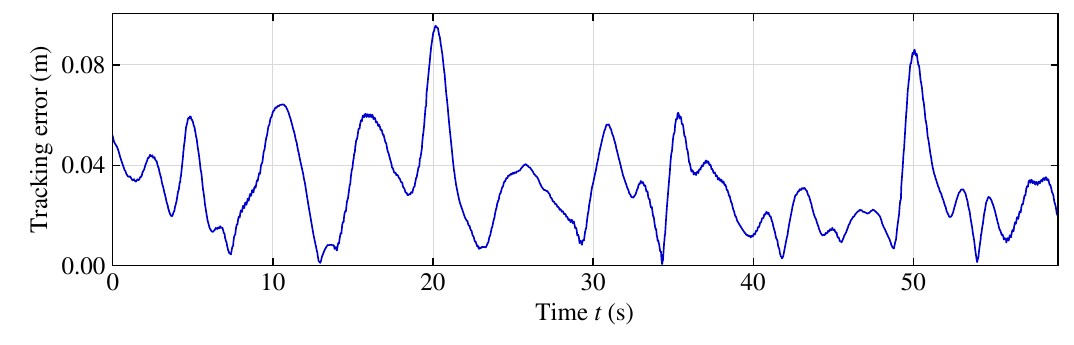}
    \caption{Curve tracking error $\sqrt{2D_{\mathcal{C}}(p(t))}$ in the experiment, in relation to the original curve $\mathcal{C}$.}
    \label{fig:track_error}
\end{figure}

The tracking performance (distance to the curve) can be seen in Figure \ref{fig:track_error}, and it is at most $0.09 \text{m}$ and $0.03 \text{m}$ on average. Figure \ref{fig:target_path} shows the agreement between the trajectory (green line) and the target curve (black).

 \begin{figure}[h!]
    \centering
    \includegraphics[trim=0 0 0 0, clip, width=1\linewidth]{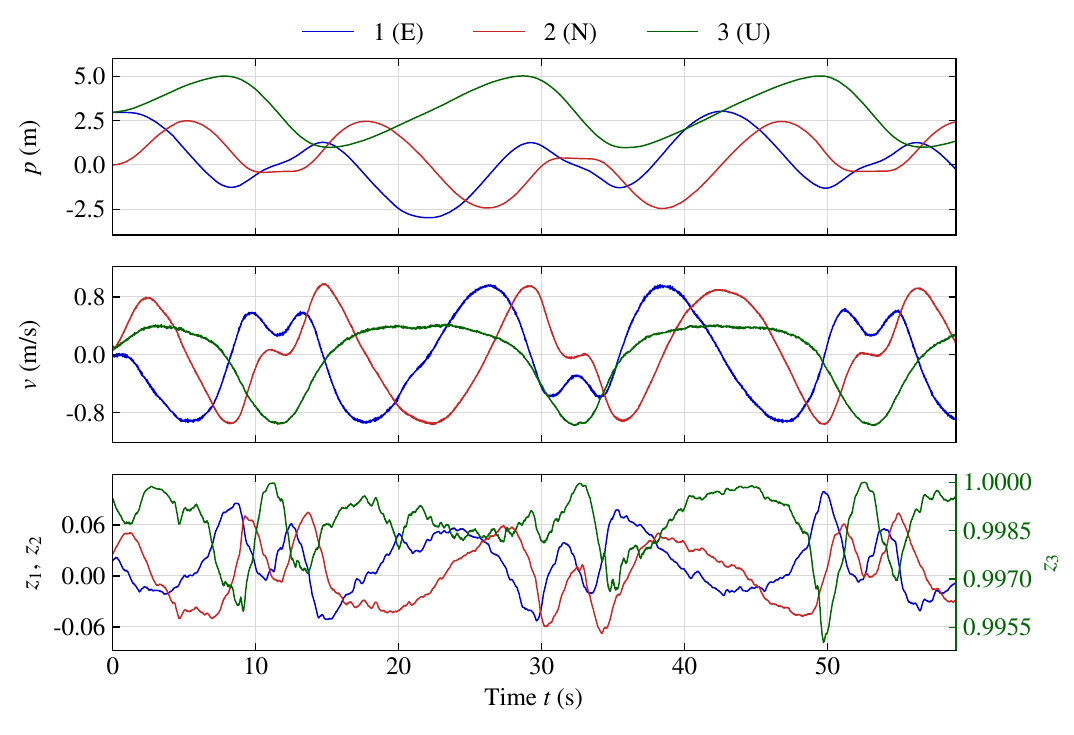}
    \caption{Physical measured states in the experiment.}
    \label{fig:ex_states}
\end{figure}

 \begin{figure}[h!]
    \centering
    \includegraphics[trim=0 0 0 0, clip, width=1\linewidth]{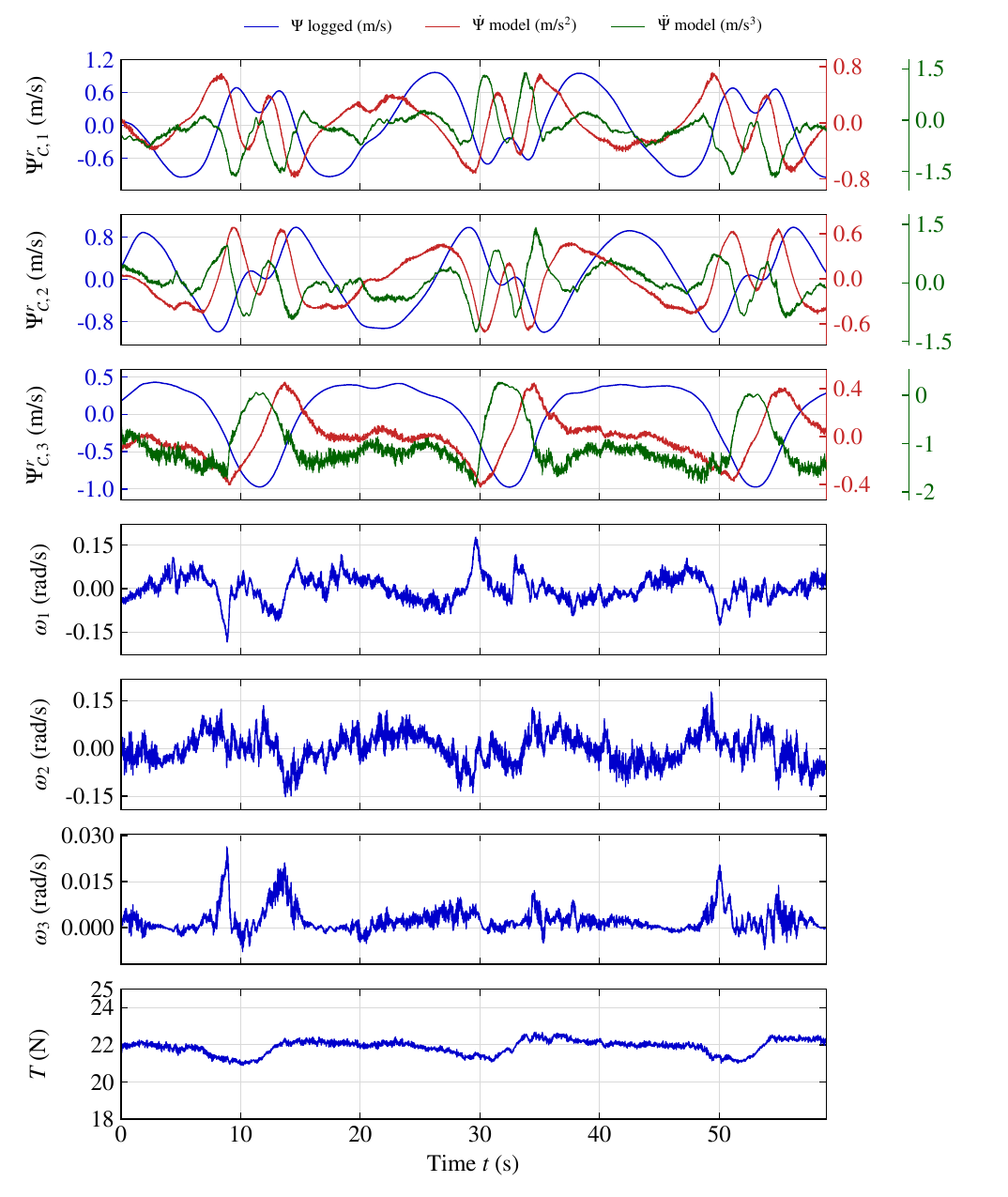}
    \caption{Control inputs in the experiment.}
    \label{fig:vf_and_controls}
\end{figure}

\begin{figure}[h!]
    \centering
    \includegraphics[trim=0 0 0 0, clip, width=1\linewidth]{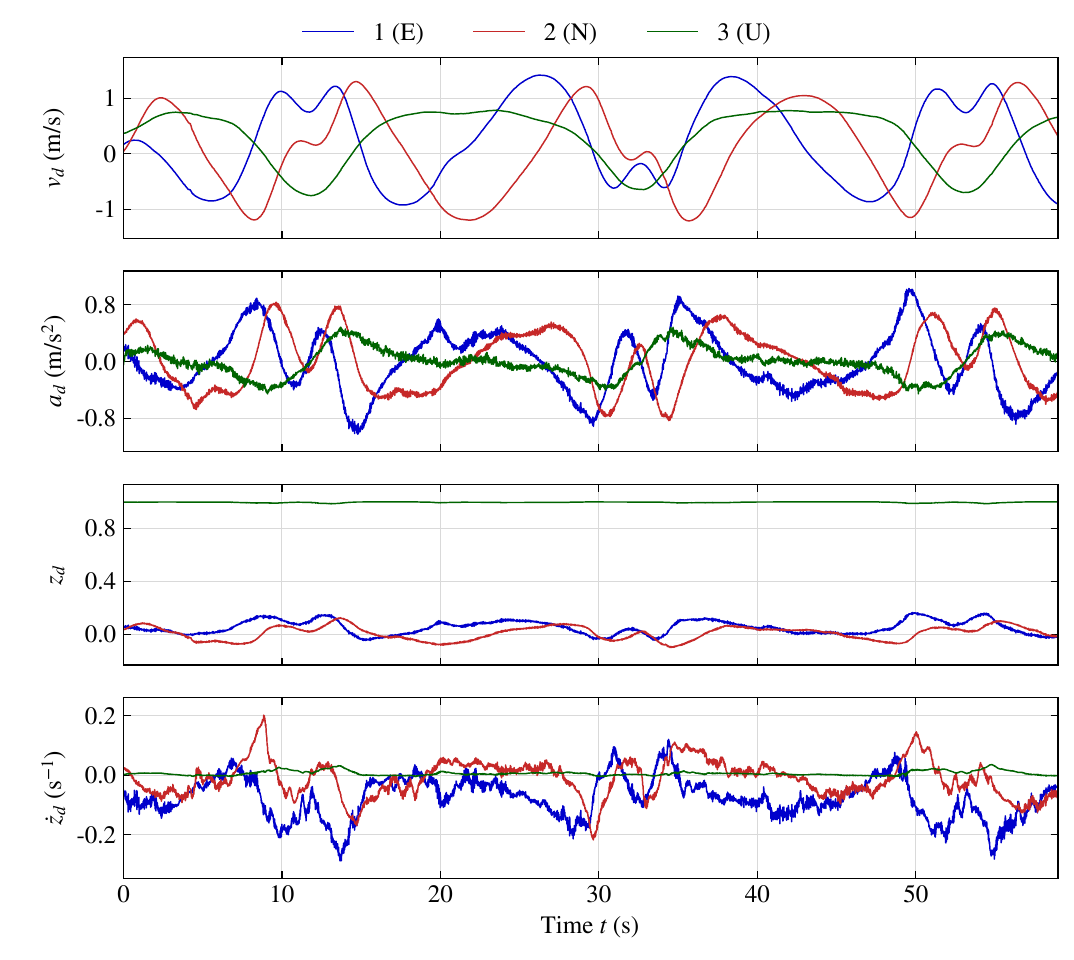}
    \caption{Auxiliary variables and virtual state $v_d$}
    \label{fig:aux}
\end{figure}

Figure \ref{fig:ex_states} shows the measured physical states. Figure \ref{fig:vf_and_controls} shows the control inputs in high level $\Psi_{\mathcal{C}}^r$, along with its first and second derivatives (needed in the controller), and the lower level control inputs $\omega$ and $T$. Figure \ref{fig:aux} shows auxiliary variables and the virtual state $s_d$ that helps understand the behavior. The high-level guidance signal $\Psi_{\mathcal{C}}^r$ and its first two derivatives vary smoothly at the scale shown. The angular velocity input $\omega$, however, exhibits more pronounced fast fluctuations. This behavior is consistent with the sensitivity of the low-level controller to fluctuations in the measured states, particularly the linear velocity $v$. Indeed, the measured velocity enters both $a_d$ and $a_c$, and its fluctuations propagate through $z_d$ and $\dot z_d$ into the angular velocity command. Spatial smoothing, as performed by our proposed  GVF, does not filter these feedback signals. Its benefit is to provide a differentiable guidance field, avoiding discontinuities associated with the curve representation and allowing the required derivatives to be computed numerically. The experiment illustrates the use of this guidance field in a controller requiring second derivatives, under real measurement conditions.

\section{Conclusion}

 In this paper, we proposed a generalization of the GVF in \cite{AdrianoIEEE} that allows a smoothing parameter $r$. For $r>0$ the GVF is infinitely differentiable, which is beneficial for control loops that require derivatives. Perhaps even more important in practice, this smoothing allows the implementation of the GVF with much less sampled points, and thus much faster, at a modest expense of tracking error.  

For future works, we would like to:

\begin{itemize}
    \item Extend to the case in which $G/H$ can be varying with $p$. Currently, $G/H$ should be constant, as opposed to \cite{AdrianoIEEE}.
    \item Extend to the time varying case, as in \cite{AdrianoIEEE}.
    \item For static curves and $r=0$, one could use structures as KD-Trees and BVH (Bounding Volume Hierarchy) to speed up the calculation from $\mathcal{O}(N)$ to $\mathcal{O}(\log(N))$. A speed up like this is probably possible for our method as well, at the expense of some numerical precision, since points $\xi$ at the curve far from $p$ contributes very little to the computation of $\Psi_{\mathcal{C}}^r(p)$. This will be studied in a  future work.
\end{itemize}
 
\section{Appendix}

To prove Proposition 1 we will need a series of lemmas. 
\begin{lemma}  \label{lemma:bound}
    Let  $\mathcal{C}$ be a regular curve and $f: \mathcal{C} \times \mathbb{R}^{n \times 1} \mapsto \mathbb{R}^{a \times b}$. Furthermore, let
    \begin{equation}
        M_{\mathcal{C}}^f(p) \triangleq \max_{\xi \in \mathcal{C}} \max_{\chi \in p^*_{\mathcal{C}}(p)} \|f(\xi,p)-f(\chi,p)\|.
    \end{equation}
    \begin{equation}
       g_{\mathcal{C}}(\xi,p) \triangleq \frac{\|\xi-p\|^2}{2}{-}D_{\mathcal{C}}(p).
    \end{equation}
    \begin{equation}
        \mathcal{G}_{\mathcal{C}}^r(p) {\triangleq} \left\{\xi {\in} \mathcal{C} \ \Big| \ g_{\mathcal{C}}(\xi,p) \le 3r^2\ln\left(\frac{\ell_{\mathcal{C}}}{2r}\right)\right\}.
    \end{equation}
    \begin{equation}
        B_{\mathcal{C}}(p) \triangleq \max_{\xi \in \mathcal{C}} \Big(1+\kappa_{\mathcal{C}}\|p-\xi\|\Big).
    \end{equation}
\begin{equation}
m^f_{\mathcal{C}}(p,r) 
\triangleq \max_{\xi \in\mathcal{G}_{\mathcal{C}}^r(p)} \max_{\chi \in p^*_{\mathcal{C}}(p)}
    \|f(\xi,p)-f(\chi,p)\|,
\end{equation}
where recall that $\ell_{\mathcal{C}}$ is the length of the curve $\mathcal{C}$.  Then, for any $0 < r< \ell_{\mathcal{C}}/2$, any  $p \in \mathbb{R}^{n \times 1}$ and any  $\chi \in p^*_{\mathcal{C}}(p)$:
    \begin{equation}
    \label{eq:deltabound}
        \Big\|\Ee[f(\xi,p)](p) - f(\chi,p) \Big\| \leq m^f_{\mathcal{C}}(p,r) + r^2M_{\mathcal{C}}^f(p)
\frac{8\sqrt{B_{\mathcal{C}}(p)}}{K\ell_{\mathcal{C}}^2},
    \end{equation}
where $ K = \int_{-1}^{1} e^{-u^2/2}du$.
\end{lemma}

\begin{proof}
\emph{Preliminary definitions:}


Fix any $p$ and any $\chi \in p_{\mathcal{C}}^*(p)$.  We start by noting that the factor $\exp(-D_{\mathcal{C}}(p))/r^2)$ cancels from
the numerator and denominator of the DEO. Thus, define
\begin{equation}
L = 
\int_{\mathcal{C}}
e^{-g_{\mathcal{C}}(\xi,p)/r^2}\|d\xi\|
\end{equation}
and $\Delta= 
\Ee[f(\xi,p)](p)-f(\chi,p)$. We have that 
\begin{equation}
\begin{aligned}
\Delta = 
\frac{1}{L}
\int_{\mathcal{C}}
e^{-g_{\mathcal{C}}(\xi,p)/r^2}
\bigl(f(\xi,p)-f(\chi,p)\bigr)\|d\xi\|.
\end{aligned}
\label{eq:master_delta_proof}
\end{equation}

We bound this quantity by lower bounding the denominator and upper bounding the numerator's norm. 

\emph{Lower bounding the denominator:}
Let $\xi(s)$ be an
arc-length parametrization such that $\xi(0)=\chi$. Define
$h(s)=g_{\mathcal{C}}\big(\xi(s),p\big)$. We have that $h(0)=0$ and, since $h(s)$ attains its minimum at $s=0$,
$\frac{d h}{ds}(0)=0$. Moreover
\begin{eqnarray}
&& \frac{d^2 h}{d s^2}(s) {=}  \Big(1{+}(\xi(s){-}p)^\top\xi''(s)\Big) \leq B_{\mathcal{C}}(p)
\end{eqnarray}
where we used the fact that, since $s$ is the arc-length, from the Frenet-Serret formulae we have $\|\xi''(s)\|\leq\kappa_{\mathcal{C}}$ for all $s$.

Note that if a twice differentiable function $k(s)$ satisfies $k''(s) \leq B$ everywhere then double integration yields $k(s) \leq k(0) + k'(0)s + Bs^2/2$. Therefore, we have that:
\begin{equation}
g_{\mathcal{C}}(\xi(s),p)=h(s)\leq\frac{B_{\mathcal{C}}(p)}{2}s^2.
\end{equation}
Assume $0<r<\ell_C/2$.
Thus, since $B_{\mathcal{C}}(p) \geq 1$, the interval
$|s|\leq r/\sqrt{B_{\mathcal{C}}(p)}$ is contained in an
arc-length interval of length $\ell_{\mathcal{C}}$. Hence,

\begin{eqnarray}
&& L
\geq
\int_{-r/\sqrt{B_{\mathcal{C}}(p)}}^{
r/\sqrt{B_{\mathcal{C}}(p)}}
e^{-B_{\mathcal{C}}(p)s^2/(2r^2)}ds = \nonumber \\
&& \frac{r}{\sqrt{B_{\mathcal{C}}(p)}}
\int_{-1}^{1}e^{-u^2/2}du  = \frac{Kr}{\sqrt{B_{\mathcal{C}}(p)}}.
\label{eq:master_denominator_bound}
\end{eqnarray}

\emph{Upper bounding the numerator's norm:}

Since $0 < r < \ell_{\mathcal{C}}/2$, we have that $3r^2 \ln(\ell_{\mathcal{C}}/(2r)) > 0$, and thus the set $\mathcal{G}_{\mathcal{C}}^r(p)$ is nonempty because it contains
$p_{\mathcal{C}}^*(p)$. From
\eqref{eq:master_delta_proof} and the triangle inequality, we upper bound the norm of the numerator of $\Delta$ by $U_1+U_2$, in which:
\begin{eqnarray}
U_1 = \int_{\mathcal{G}_{\mathcal{C}}^r(p)}
e^{-g_{\mathcal{C}}(\xi,p)/r^2}
\|f(\xi,p)-f(\chi,p)\|\|d\xi\|. \nonumber \\
U_2 = 
\int_{\mathcal{C}-\mathcal{G}_{\mathcal{C}}^r(p)}
e^{-g_{\mathcal{C}}(\xi,p)/r^2}
\|f(\xi,p)-f(\chi,p)\|\|d\xi\|.
\label{eq:master_split}
\end{eqnarray}

By the definition of $m_{\mathcal{C}}^f(p,r)$,  $U_1  \leq m_{\mathcal{C}}^f(p,r)L$. For $U_2$, we note that, on $\mathcal{C}-\mathcal{G}_{\mathcal{C}}^r(p)$,
\begin{equation}
\begin{aligned}
e^{-g_{\mathcal{C}}(\xi,p)/r^2}
&\leq
e^{-\big(3r^2\ln (\ell_{\mathcal{C}}/(2r))\big)/(r^2)}
=
\left(\frac{2r}{\ell_{\mathcal{C}}}\right)^3.
\end{aligned}
\end{equation}
Therefore, $U_2 \leq \ell_{\mathcal{C}} \cdot M_{\mathcal{C}}^f(p)(8r^3/ \ell_{\mathcal{C}}^3)$.

The quantity $\|\Delta\|$ is thus upper bounded by $U_1/L + U_2/L$. The first term is bounded by $m_{\mathcal{C}}^f(p,r)$, while  the second term satisfies 
\begin{equation}
\begin{aligned}
\frac
{8M_{\mathcal{C}}^f(p)r^3
}{
L\ell_{\mathcal{C}}^2
}
&\leq
\frac{8M_{\mathcal{C}}^f(p)r^3
\sqrt{B_{\mathcal{C}}(p)}
}{
 Kr\ell_{\mathcal{C}}^2
}
\
&=
r^2M_{\mathcal{C}}^f(p)
\frac{8\sqrt{B_{\mathcal{C}}(p)}}{K\ell_{\mathcal{C}}^2}.
\end{aligned}
\end{equation}
And the result in \eqref{eq:deltabound} is proved. $\blacksquare$
\end{proof}

This lemma stablishes bounds, but we now show that for some points $p$ and assuming continuity of $f$, this bound can be made arbitrarily small by making $r$ small.

\begin{lemma} \label{lemma:convEe}
    Let  $\mathcal{C}$ be a regular curve and $f: \mathcal{C} \times \mathbb{R}^{n \times 1} \mapsto \mathbb{R}^{a \times b}$ a continuous function in both arguments.  Define the subset of $\mathbb{R}^{n \times 1}$ for all points $p$ in which multiple values of $p^*_{\mathcal{C}}(p)$ does not create a multi-valued $f$ \footnote{This is a generalization of $\mathcal{U}_{\mathcal{C}}$, since $\mathcal{U}_{\mathcal{C}} = \mathcal{U}^f_{\mathcal{C}}$ when $f(\xi,p) = \xi$. Furthermore, note that $\mathcal{U}_{\mathcal{C}} \subseteq \mathcal{U}^f_{\mathcal{C}}$ for all $f$.}:
    \begin{equation}
        \mathcal{U}_{\mathcal{C}}^f \triangleq \{p \in \mathbb{R}^{n \times 1} \ | \ f(\xi_1,p) = f(\xi_2,p) \ \forall \xi_1, \xi_2 \in p^*_{\mathcal{C}}(p)\}.
    \end{equation}

        Given a point $p \in \mathcal{U}_{\mathcal{C}}^f$, denote the common evaluation of $f$ at $p$ by $f^*(p)$.  Then, when $r \rightarrow +0$, $\Ee[f(\xi,p)](p) \rightarrow f^*(p)$.
\end{lemma}
\begin{proof}
We invoke Lemma \ref{lemma:bound} and analyze the behavior of \eqref{eq:deltabound}. Note that for a fixed $p \in \mathcal{U}_{\mathcal{C}}^f$, $\lim_{r \rightarrow +0} m_{\mathcal{C}}^f(p,r) = 0$, because $g_{\mathcal{C}}(\xi,p) \geq 0$ and  $g_{\mathcal{C}}(\xi,p) = 0$ implies $\xi \in p^*_{\mathcal{C}}(p)$, and there, for any such $\xi$, $f(\xi,p) = f^*(p)$. This and the continuity of $f$ guarantee that $\lim_{r \rightarrow +0} m^{f}_{\mathcal{C}}(p,r) = 0$. Clearly, the other term also vanishes when $r \rightarrow +0$.
$\blacksquare$
\end{proof}

As we get closer and closer to a point $p \not \in \mathcal{U}_{\mathcal{C}}^f$, the convergence becomes slower, because $m_{\mathcal{C}}^f$ falls to $0$ slower. 

\begin{lemma} \label{lemma:derivative} Let $\mathcal{C}$ be a regular curve and $f: \mathcal{C} \times \mathbb{R}^{n \times 1} \mapsto \mathbb{R}^{m \times 1}$ be a function that is differentiable in the second variable. Define  $\text{Cov}_{\mathcal{C}}^r[f(\xi,p),g(\xi,p)](p)$ as the covariance matrix between $f(\xi,p)$ and another function $g: \mathcal{C} \times \mathbb{R}^{n \times 1} \mapsto \mathbb{R}^{m \times 1}$ :

\begin{eqnarray}
 \text{Cov}_{\mathcal{C}}^r[f(\xi,p),g(\xi,p)] \triangleq  && \Ee\big[ f(\xi,p)g(\xi,p)^{\top}\big]\nonumber \\ 
 -&& \Ee\big[ f(\xi,p)\big]\Ee\big[ g(\xi,p)\big]^{\top}.
\end{eqnarray}
Then:
\begin{eqnarray}
    && \frac{\partial}{\partial p}\Ee[f(\xi,p)] {=}\Ee\left[ \frac{\partial f}{\partial p}(\xi,p)\right] {+} \frac{\text{Cov}_{\mathcal{C}}^r[f(\xi,p),\xi-p]}{r^2}.\nonumber \\
\end{eqnarray}

\end{lemma}
\begin{proof} By inspection. $\blacksquare$.
    
\end{proof}

\begin{lemma}  \label{lemma:divbound} Let $\mathcal{C}$ be a regular curve. Define the \emph{$r$-DEO-variance} of a function $f: \mathcal{C} \times \mathbb{R}^{n \times 1} \mapsto \mathbb{R}^{m \times 1}$:
\begin{equation}
    \text{Var}_{\mathcal{C}}^r\big[f(\xi,p)\big] \triangleq \Ee\big[\|f(\xi,p)\|^2\big]-\big\|\Ee[f(\xi,p)]\big\|^2 .
\end{equation}

Consider $\Psi_{\mathcal{C}}^{r}(p)$ as in \eqref{eq:PsiCr}, with $p \in \mathbb{R}^{n \times 1}$. Then:
\begin{eqnarray}
    \text{Tr}\left(\frac{\partial \Psi_{\mathcal{C}}^r}{\partial p}(p) \right) = && G \left(\frac{\text{Var}_{\mathcal{C}}^r\big[\xi-p\big]}{r^2}-n\right) \nonumber \\
    && -\frac{H}{r^2}\Ee[T_{\mathcal{C}}(\xi)]^{\top}\Ee[\xi-p].
\end{eqnarray}
\end{lemma}
\begin{proof}

Take $f(\xi,p) = \psi_{\mathcal{C}}(\xi,p) = G(\xi-p)+HT_{\mathcal{C}}(\xi)$.  Thus, using Lemma \ref{lemma:derivative}, the Jacobian of $\Psi^r_{\mathcal{C}}(p) = \Ee[\psi_{\mathcal{C}}(\xi,p) ](p)$ is:
\begin{eqnarray}
     && G\left(\frac{\text{Cov}_{\mathcal{C}}^r[\xi-p,\xi-p]}{r^2}{-}I_{n \times n}\right) \nonumber\\ 
     &&-\frac{H}{r^2}\Ee[T_{\mathcal{C}}(\xi)]\Ee[\xi-p]^{\top}\nonumber \\
     && +Hr^{-2} \Ee\big[T_{\mathcal{C}}(\xi)(\xi-p)^{\top}\big].
\end{eqnarray}
Now, take the trace of this matrix, use the linearity of $\Ee$ and the fact that $\text{Tr}(ab^{\top}) = b^{\top}a$ for two column vectors $a,b$, so we can conclude that the trace of the Jacobian is:

\begin{eqnarray}
  &&G\left(\frac{\text{Var}_{\mathcal{C}}^r[\xi-p]}{r^2}{-}n\right)-\frac{H}{r^2}\Ee[T_{\mathcal{C}}(\xi)]^{\top}\Ee[\xi-p]\ \nonumber\\
  &&+Hr^{-2}\Ee\big[(\xi-p)^{\top}T_{\mathcal{C}}(\xi)\big].
\end{eqnarray}
The last term can be shown to be zero. This is because:

\begin{eqnarray}
    && \int_{\mathcal{C}}e^{-\frac{\|p-\xi\|^2}{2r^2}}(\xi-p)^{\top}\underbrace{T_{\mathcal{C}}(\xi)\|d\xi\|}_{d\xi} = \nonumber \\
   && -r^2\int_{\mathcal{C}} \frac{\partial}{\partial \xi}\left( e^{-\frac{\|p-\xi\|^2}{2r^2}}\right)d\xi = 0
\end{eqnarray}

\noindent since the curve $\mathcal{C}$ is closed. This concludes the proof. $\blacksquare$
\end{proof}

\begin{lemma}
\label{lemma:divbound2}
     Let $\mathcal{C}$ be a regular curve. Define:
     \begin{equation}
     \label{eq:Zdef}
         Z_{\mathcal{C}}^r(p) \triangleq \int_{\mathcal{C}} e^{-\frac{\|\xi-p\|^2}{2r^2}}\|d\xi\|.
     \end{equation}
     Also, define the set:
     \begin{equation}
     \label{eq:Omegadef}
         \Omega_{\mathcal{C}}^r \triangleq \{p \in \mathbb{R}^{n \times 1} \ | \ D_{\mathcal{C}}(p) \geq nr^2\}.
     \end{equation} 
     Let $\tilde{\Psi}_{\mathcal{C}}^r(p) \triangleq  Z_{\mathcal{C}}^r(p) \Psi_{\mathcal{C}}^r(p)$, with $\Psi_{\mathcal{C}}^{r}(p)$ as in \eqref{eq:PsiCr}. Then, for all $p \in  \Omega_{\mathcal{C}}^r$:
     \begin{equation}
          \text{Tr}\left(\frac{\partial \tilde{\Psi}_{\mathcal{C}}^r}{\partial p}(p) \right) \geq G nZ_{\mathcal{C}}^r(p).
     \end{equation}
     
\end{lemma}

\begin{proof}
Using standard calculus rules and the properties of the trace operator, we have that:
\begin{equation}
 \text{Tr}\left(\frac{\partial \tilde{\Psi}_{\mathcal{C}}^r}{\partial p}(p) \right) = Z_{\mathcal{C}}^r(p) \text{Tr}\left(\frac{\partial \Psi_{\mathcal{C}}^r}{\partial p}(p) \right)  + \frac{\partial Z_{\mathcal{C}}^r}{\partial p}(p)\Psi_{\mathcal{C}}^r(p).
\end{equation}
By noting that $\frac{\partial Z_{\mathcal{C}}^r}{\partial p}(p) = \frac{Z_{\mathcal{C}}^r(p)}{r^2}\Ee[\xi-p]^{\top}$, and using the result in Lemma \ref{lemma:divbound}, we obtain that:

\begin{equation}
 \text{Tr}\left(\frac{\partial \tilde{\Psi}_{\mathcal{C}}^r}{\partial p}(p) \right) = GZ_{\mathcal{C}}^r(p)\left(\frac{\Ee[\|\xi-p\|^2]}{r^2}{-}n\right).
\end{equation}

Note that, since $\|\xi-p\|^2 \geq 2D_{\mathcal{C}}(p)$ for $\xi \in \mathcal{C}$ and $\Ee$ is a monotone operator, $\Ee[\|\xi-p\|^2] \geq  2D_{\mathcal{C}}(p)$. Thus, for $p \in \Omega_{\mathcal{C}}^r $, it is clear that the desired result holds. $\blacksquare$
\end{proof}

\begin{lemma} \label{lemma:convergence} Let $\mathcal{C}$ be a regular curve and consider the dynamical system $\dot{p} = \Psi_{\mathcal{C}}^{r}(p)$ with $p \in \mathbb{R}^{n \times 1}$ and $\Psi_{\mathcal{C}}^{r}(p)$ as in \eqref{eq:PsiCr}. Then, the set of initial conditions $p_0$ that stays in the set $\Omega_{\mathcal{C}}^r$  (see \eqref{eq:Omegadef}) has measure zero in $\mathbb{R}^{n \times 1}$.
    
\end{lemma}
\begin{proof}
    The key point is to define a weighted measure in $\mathbb{R}^{n \times 1}$ instead of the traditional one. Let $\text{Vol}_{\mathcal{C}}^r: 2^{\mathbb{R}^{n \times 1}} \mapsto \mathbb{R}^+$ be defined as:
    \begin{equation}
        \text{Vol}_{\mathcal{C}}^r(\mathcal{A}) = \int_{\mathcal{A}} Z_{\mathcal{C}}^r(p)dp
    \end{equation}
    \noindent with $Z_{\mathcal{C}}^r(p)$ as in \eqref{eq:Zdef}. It is finite for all sets, indeed:
    \begin{equation}
       \text{Vol}_{\mathcal{C}}^r(\mathbb{R}^{n \times 1}) \leq (2\pi)^{n/2}r^n \ell_{\mathcal{C}} 
    \end{equation}
    \noindent in which the $n$-dimensional Gaussian integral was employed. 

    Let $\Phi(t,p_0)$ be the flow of the differential equation $\dot{p} = \Psi_{\mathcal{C}}^r(p)$. Now consider the set $ \mathcal{A} = \{p_0 \in \mathbb{R}^{n \times 1} \ | \ D_{\mathcal{C}}\big(\Phi(t,p_0)\big) \geq nr^2 \ \forall t \geq 0\}$.  We will show that this set has zero measure in $\mathbb{R}^{n \times 1}$.

    Using $\tilde{\Psi}_{\mathcal{C}}^r(p) = Z_{\mathcal{C}}^r(p)\Psi_{\mathcal{C}}^r(p)$  We use the weighted Liouville Formula \cite{Rantzer2001}:
    \begin{eqnarray}
       && Z_{\mathcal{C}}^r\big(\Phi(t,p_0)\big)\text{det}\left(\frac{\partial \Phi}{\partial p_0}(t,p_0)\right) = \nonumber\\
       && Z_{\mathcal{C}}^r(p_0)\exp\left(\int_0^t \frac{1}{Z_{\mathcal{C}}^r\big(\Phi(s,p_0)\big)}\text{Tr}\left(\frac{\partial \tilde{\Psi}_{\mathcal{C}}^r}{\partial p}\big(\Phi(s,p_0)\big) \right)ds \right) \nonumber \\
    \end{eqnarray}
    Now take a $p_0 \in \mathcal{A} \subseteq \Omega_{\mathcal{C}}^r$ (see \eqref{eq:Omegadef}). There, we can use Lemma \ref{lemma:divbound2}  to conclude that:
    \begin{equation}
    \label{eq:Zbound}
       Z_{\mathcal{C}}^r\big(\Phi(t,p_0)\big)\text{det}\left(\frac{\partial \Phi}{\partial p_0}(t,p_0)\right)  \geq Z_{\mathcal{C}}^r(p_0) e^{Gn t}. 
    \end{equation}
    Now, we use the change of variable formula and \eqref{eq:Zbound} to conclude  that:
    \begin{eqnarray}
       &&  \text{Vol}_{\mathcal{C}}^r\big(\Phi(t,\mathcal{A})\big) = \int_{\mathcal{A}} Z_{\mathcal{C}}^r(\Phi(t,p_0))d\Phi(t,p_0) = \nonumber \\ 
       && \int_{\mathcal{A}} Z_{\mathcal{C}}^r(\Phi(t,p_0))\text{det}\left(\frac{\partial \Phi}{\partial p_0}(t,p_0)\right)dp_0 \nonumber \\ 
       && \underbrace{\geq}_{\mbox{eq. \eqref{eq:Zbound}}}   e^{Gn t} \text{Vol}_{\mathcal{C}}^r\big(\mathcal{A}\big).
    \end{eqnarray}

    Thus, for all $t \geq 0$, $\text{Vol}_{\mathcal{C}}^r\big(\Phi(t,\mathcal{A})\big) \geq  e^{Gn t} \text{Vol}_{\mathcal{C}}^r\big(\mathcal{A}\big)$. 
    
    Now, we note that from triangle inequality $\|\Psi_{\mathcal{C}}^r(p)\| \leq G \|p\| + G \max_{\xi \in \mathcal{C}} \|\xi\| + H$. Thus $\|\Psi_{\mathcal{C}}^r(p)\|$ has a linearly-bounded growth and thus the vector field is forward complete, i.e., $\Phi(t,p_0)$ is finite for all finite $t,p_0$. Thus, if we make $t \rightarrow \infty$, and use the fact that $Gn > 0$ and that $\text{Vol}_{\mathcal{C}}^r\big(\Phi(t,\mathcal{A})\big)$ is bounded, we find that $\text{Vol}_{\mathcal{C}}^r\big(\mathcal{A}\big) = 0$. Thus, it has zero measure in $\mathbb{R}^{n \times 1}$, since $Z_{\mathcal{C}}^r(p) > 0$. 
    
    Hence, any set of points that $D_{\mathcal{C}}(p) \geq nr^2$ forever has zero measure in $\mathbb{R}^{n \times 1}$. And the proof is concluded. $\blacksquare$
\end{proof}

\begin{lemma} \label{lemma:persist} Let $\mathcal{C}$ be a regular curve and consider the dynamical system $\dot{p} = \Psi_{\mathcal{C}}^{r}(p)$ with $p \in \mathbb{R}^{n \times 1}$ and $\Psi_{\mathcal{C}}^{r}(p)$ as in \eqref{eq:PsiCr}. Then, for $r$ small enough, there is a compact set $\mathcal{C}^r $ that is positively invariant. Furthermore, $\mathbb{R}^{n \times 1} - \Omega_{\mathcal{C}}^r \subseteq \mathcal{C}^r $, $\mathcal{C}^r \subseteq \mathcal{U}_{\mathcal{C}}$ and $\lim_{r \rightarrow +0} \mathcal{C}^r  = \mathcal{C}$.
    
\end{lemma}

\begin{proof}
Let $ \delta_{\mathcal{C}} = \inf_{p \not \in \mathcal{U}_{\mathcal{C}}} D_{\mathcal{C}}(p)$. By the tubular-neighborhood theorem \cite{Foote1984} for compact embedded
twice differentiable curves (which is the case of regular curves), there exists a neighborhood of $\mathcal{C}$
in which every point has a unique closest point on
$\mathcal{C}$. Consequently, $\delta_{\mathcal{C}}>0$. 

Now, let $\hat{\mathcal{C}} = \{p \in \mathbb{R}^{n \times 1} \ | \ D_{\mathcal{C}}(p)\leq \frac{\delta_{\mathcal{C}}}{2}\} \subseteq \mathcal{U}_{\mathcal{C}}$. Consider a  $p \in \hat{\mathcal{C}}$, and thus $p \in \mathcal{U}_{\mathcal{C}}$. In $\mathcal{U}_{\mathcal{C}}$, we can make $r$ small as possible (Lemma \ref{lemma:convEe}) so $f(\xi,p) \rightarrow f(p^*_{\mathcal{C}}(p),p)$ for any continuous function $f$. Furthermore, since the set $\hat{\mathcal{C}}$ is compact, we can make this bound independent of $p$. 

Let $e_N(p) = \Ee[\xi{-}p]-\big(p^*_{\mathcal{C}}(p)-p\big)$ and $e_T(p)  = \Ee[T_{\mathcal{C}}(\xi)]-T_{\mathcal{C}}\big(p^*_{\mathcal{C}}(p)\big)$.
Thus, we can have functions $\epsilon_N(r), \epsilon_T(r): \mathbb{R}^+ \mapsto \mathbb{R}^+$ that decrease to $0$ as $r \rightarrow +0$. Therefore, for all $p \in \hat{\mathcal{C}}$:
\begin{equation}
    \|e_N(p)\| \leq \epsilon_N(r) \ , \ \|e_T(p)\| \leq \epsilon_T(r).
\end{equation}
Let $\epsilon(r) =  \epsilon_N(r)+\frac{H}{G} \epsilon_T(r)$ and $c(r) = \max(nr^2,\frac{1}{2}\epsilon(r)^2)$. Pick $r$ small enough so $c(r) \leq \frac{1}{2}\delta_{C}$ and define 
\begin{equation}
    \mathcal{C}^r \triangleq \{ p \in \mathbb{R}^{n \times 1} \ | \ D_{\mathcal{C}}(p) \leq c(r)\}
\end{equation}
\noindent and thus $\mathcal{C}^r \subseteq \mathcal{U}_{\mathcal{C}}$. Consider now that  $p \in  \mathcal{C}^r$.

Thus, using the fact that $\Psi_{\mathcal{C}}^r(p) = G(\big(p^*_{\mathcal{C}}(p)-p\big)+e_N) + H(T_{\mathcal{C}}\big(p^*_{\mathcal{C}}(p)\big)+e_T)$, the fact that $\big(p-p^*_{\mathcal{C}}(p)\big)^{\top} T_{\mathcal{C}}\big(p^*_{\mathcal{C}}(p)\big) = 0$ and Schwartz's inequality:
\begin{equation}
  \frac{d}{dt} D_{\mathcal{C}}(p) \leq -2GD_{\mathcal{C}}(p) +G\sqrt{2D_{\mathcal{C}}(p)}\epsilon(r)
\end{equation}
\noindent and one can check $\frac{d}{dt} D_{\mathcal{C}}(p) \leq 0$ at the boundary of $\mathcal{C}^r$. This implies that $\mathcal{C}^r$ is positively invariant. This set is contained in $\mathcal{U}_{\mathcal{C}}$ and is compact by construction. Furthermore, since $c(r) \geq nr^2$, we have that $\mathcal{C}^r \supseteq \mathbb{R}^{n \times 1}-\Omega_{\mathcal{C}}^r$. In addition, by the definition of $\mathcal{C}^r $, clearly $\lim_{r \rightarrow +0} \mathcal{C}^r  = \mathcal{C}$. $\blacksquare$
\end{proof}

\begin{lemma} \label{lemma:neverstop} Let $\mathcal{C}$ be a regular curve and consider the dynamical system $\dot{p} = \Psi_{\mathcal{C}}^{r}(p)$ with $p \in \mathbb{R}^{n \times 1}$ and $\Psi_{\mathcal{C}}^{r}(p)$ as in \eqref{eq:PsiCr}. Let $\mathcal{A} \subseteq \mathcal{U}_{\mathcal{C}}$ be a compact subset. For $r$ small enough, no point of $\mathcal{A}$ can be such that $\Psi_{\mathcal{C}}^{r}(p)=0$.
    
\end{lemma}
\begin{proof}
The proof follows by contradiction. Suppose that there exists a compact $\mathcal{A} \subseteq \mathcal{U}_{\mathcal{C}}$ and that there exists $p \in \mathcal{A}$ such that $\Psi_{\mathcal{C}}^r(p) = 0$.

Applying Lemma \ref{lemma:convEe} with $f=\psi_{\mathcal{C}}$, since $\mathcal{A}$ is compact, we can have $r$ small enough so, for all $p \in \mathcal{A}$, we have
    \begin{equation}
    \label{eq:Hbound}
        \|\underbrace{\Psi_{\mathcal{C}}^r(p)}_{=0} - \Psi_{\mathcal{C}}(p)\| < H
    \end{equation}
    \noindent in which $H>0$. Thus, $\|\Psi_{\mathcal{C}}(p)\| < H$. But this is impossible, because $\|\Psi_{\mathcal{C}}(p)\| = \sqrt{2G^2D_{\mathcal{C}}(p)+H^2} \geq H$,  which contradicts \eqref{eq:Hbound}.
    $\blacksquare$
\end{proof}
Finally, we can prove \textbf{Proposition 1}:

\textbf{First statement}: comes trivially from the definition of $\Psi_{\mathcal{C}}^r(p)$.

\textbf{Second statement}: comes from Lemma \ref{lemma:convEe}, by using the function $f(\xi,p) = \psi_{\mathcal{C}}(\xi,p)$.

\textbf{Third statement}:  from Lemma \ref{lemma:convergence}, for almost all initial conditions, we cannot stay inside  $\Omega_{\mathcal{C}}^r$ forever. Thus we must enter, in finite time, $\mathbb{R}^{n \times 1}-\Omega_{\mathcal{C}}^r$. Using Lemma \ref{lemma:persist}, if $r$ is small enough, this implies that we entered $\mathcal{C}^r$, which is positively invariant, and thus we never leave this set.  Furthermore, this set $\mathcal{C}^r$ is such that $\lim_{r \rightarrow +0} \mathcal{C}^r = \mathcal{C}$. Finally, since $\mathcal{C}^r \subseteq \mathcal{U}_{\mathcal{C}}$ and it is compact, Lemma \ref{lemma:neverstop} says that the vector field never vanishes there. Thus, the point stay moving, with a vector field which is an approximation of the tangent field because, for $r$ small inside $\mathcal{C}^r$, $\Ee[\xi-p](p) \approx 0$ and $\Ee[T_{\mathcal{C}}(\xi)](p) \approx T_{\mathcal{C}}\big(p^*_{\mathcal{C}}(p)\big)$.

\bibliographystyle{ieeetr}  
\bibliography{bibl}
\end{document}